\documentclass[10pt,twocolumn,letterpaper]{article}

\usepackage[pagenumbers]{cvpr} 
\usepackage{graphicx}
\usepackage{amsmath}
\usepackage{amssymb}
\usepackage{mathtools}
\usepackage{booktabs}
\usepackage{my_style}
\usepackage{cuted}

\usepackage[detect-all,per-mode=symbol,group-digits=integer,group-minimum-digits=4]{siunitx}

\let\norm\relax
\DeclarePairedDelimiterX{\norm}[1]{\lVert}{\rVert}{#1}

\AtBeginEnvironment{tabular}{\normalsize}

\usepackage{hyperref}
\hypersetup{%
  colorlinks=true,
  urlbordercolor={1 1 1},
  linkcolor={black},
  raiselinks=false
}

\usepackage[capitalize]{cleveref}
\crefname{section}{Sec.}{Secs.}
\Crefname{section}{Section}{Sections}
\Crefname{table}{Table}{Tables}
\crefname{table}{Tab.}{Tabs.}

\begin{document}

\title{DKM: Dense Kernelized Feature Matching for Geometry Estimation}

\author{\large{Johan Edstedt, \quad Ioannis Athanasiadis, \quad Mårten Wadenbäck, \quad Michael Felsberg}\\
\normalsize{Computer Vision Laboratory}\\
\normalsize{Linköping University}\\
{\tt\small \{firstname\}.\{lastname\}@liu.se}}
\maketitle

\begin{abstract}
Feature matching is a challenging computer vision task that involves finding correspondences between two images of a 3D scene. 
In this paper we consider the dense approach instead of the more common sparse paradigm, thus striving to find all correspondences. 
Perhaps counter-intuitively, dense methods have previously shown inferior performance to their sparse and semi-sparse counterparts for estimation of two-view geometry.
This changes with our novel dense method, which outperforms both dense and sparse methods on geometry estimation. The novelty is threefold: First, we propose a kernel regression global matcher. Secondly, we propose warp refinement through stacked feature maps and depthwise convolution kernels. Thirdly, we propose learning dense confidence through consistent depth and a balanced sampling approach for dense confidence maps.

Through extensive experiments we confirm that our proposed dense method, \textbf{D}ense \textbf{K}ernelized Feature \textbf{M}atching, sets a new state-of-the-art on multiple geometry estimation benchmarks. In particular, we achieve an improvement on MegaDepth-1500 of +4.9 and +8.9 AUC$@5^{\circ}$ compared to the best previous sparse method and dense method respectively. Our code is provided at the following repository: \url{https://github.com/Parskatt/dkm}.
\end{abstract}

\section{Introduction}
\begin{figure}[ht!]
    \centering
    \includegraphics[width=\linewidth]{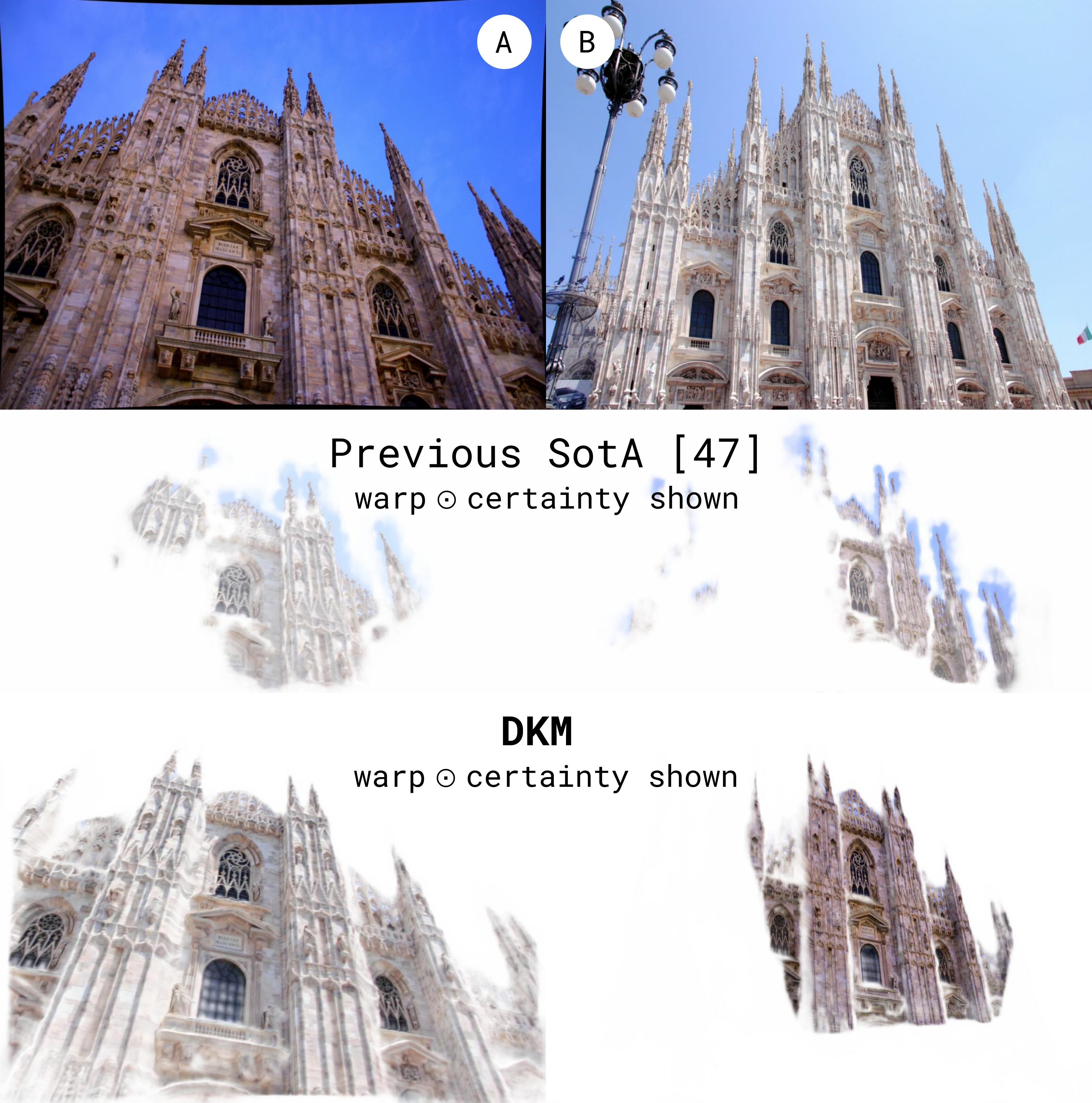}
    \caption{Comparison between our proposed approach \textbf{DKM} and the previous SOTA method PDC-Net+~\cite{truong2021pdc} on Milan Cathedral. Top row, image \A~and \B. Middle row and bottom row, forward and reverse warps for PDC-Net+ and DKM weighted by certainty. DKM provides both superior match accuracy and certainty estimation compared to previous methods.}
    
    \label{fig:qualitative}
\end{figure}
Two-view geometry estimation is a classical computer vision problem with numerous important applications, including 3D reconstruction~\cite{schonberger2016structure}, SLAM~\cite{mur2015orb}, and visual re-localisation~\cite{lynen2020large}. The task can roughly be divided into two steps. First, a set of matching pixel pairs between the images is produced. Then, using the matched pairs, two-view geometry,~\eg, relative pose, is estimated. In this paper, we focus on the first step,~\ie, feature matching. This task is challenging, as image pairs may exhibit extreme variations in viewpoint~\cite{li2018megadepth}, illumination~\cite{balntas2017hpatches}, time of day~\cite{sattler2018benchmarking}, and even season~\cite{toft2020long}. This stands in contrast to small baseline stereo and optical flow tasks, where the changes in viewpoint and illumination are typically small.

Traditionally, feature matching has been performed by sparse keypoint and descriptor extraction, followed by matching~\cite{lowe2004distinctive,sarlin2020superglue}. The main issue with this approach is that accurate localization of reliable and repeatable keypoints is difficult in challenging scenes. This leads to errors in matching and estimation~\cite{germain2020s2dnet,lindenberger2021pixel}. To tackle this issue, semi-sparse or \emph{detector-free} methods such as LoFTR~\cite{sun2021loftr} and Patch2Pix~\cite{zhou2021patch2pix} were introduced. These methods do not detect keypoints directly but rather perform global matching at a coarse level, followed by mutual nearest neighbour extraction and sparse match refinement. While those methods degrade less in low-texture scenes, they are still limited by the fact that the sparse matches are produced at a coarse scale, leading to problems with,~\eg, repeatability due to grid artifacts~\cite{hetech}.
By instead extracting \emph{all} matches between the views,~\ie, \emph{dense} feature matching, we face no such issues. Furthermore, dense warps provide affine matches for free, which yield smaller minimal problems for subsequent estimation~\cite{barath2017minimal,barath2020making,guan2020minimal}. While previous dense approaches~\cite{shen2020ransac,truong2020glu} have achieved good results, they have however failed to achieve performance rivaling that of sparse or semi-sparse methods on geometry estimation.

In this work, we propose a novel dense matching method that outperforms both dense and sparse methods in homography and two-view relative pose estimation. We achieve this by proposing a substantially improved model architecture, including both the global matching and warp refinement stage, and by a simple but strong approach to dense certainty estimation and a balanced dense warp sampling mechanism. We compare qualitatively our method with the previous best dense method in Figure~\ref{fig:qualitative}.

Our \textbf{contributions} are as follows.~\textbf{Global Matcher:} We propose a kernelized global matcher and embedding decoder. This results in robust coarse matches. We describe our approach in Section~\ref{sec:global_matcher} and ablate the performance gains in Table~\ref{tab:abl_global_matcher}. \textbf{Warp Refiners:} We propose warp refinement through large depthwise separable kernels using stacked feature maps as well as local correlation as input. This gives our method superior precision and is described in detail in Section~\ref{sec:refiner} with corresponding performance impact ablated in Table~\ref{tab:abl_refiners}. \textbf{Certainty and Sampling:} We propose a simple method to predict dense certainty from consistent depth and propose a balanced sampling approach for dense matches. We describe our certainty and sampling approach in more detail in Section~\ref{sec:sampling} and ablate the performance gains in Table~\ref{tab:abl_sampling}.
\textbf{State-of-the-Art:} Our extensive experiments in Section~\ref{sec:sota} show that our method significantly improves on the state-of-the-art. In particular, we improve estimation results compared to the best previous dense method by +8.9 AUC$@5^{\circ}$ on MegaDepth-1500. These results pave the way for dense matching based 3D reconstruction.

\section{Related Work}
\parsection{Global Matching}
Traditionally, global matching has been performed by computing pair-wise descriptor distances for detected keypoints in the two images, with match extraction performed by mutual nearest neighbours in the distance matrix, see \eg~\cite{lowe2004distinctive,detone2018superpoint,dusmanu2019d2}.
Instead of directly computing pair-wise distances, one can first condition the descriptors based on the complete set of detections. Sarlin~\etal~\cite{sarlin2020superglue} proposed a graph neural network approach to condition the descriptors, and optimal transport instead of mutual nearest neighbours for match extraction. 
Detector-free methods instead perform global matching uniformly over the image grid at a coarse scale~\cite{rocco2018neighbourhood,rocco2020efficient,tinchev2020xrcnet, zhou2021patch2pix}. This has the benefit of avoiding the detection problem~\cite{sun2021loftr}. These methods typically extract matches by (soft-)mutual-nearest neighbours, or optimal transport~\cite{rocco2018neighbourhood,sun2021loftr}. In contrast to detector-free methods, dense methods must produce a dense warp. This warp is typically predicted by regression based on the global 4D-correlation volume~\cite{melekhov2019dgc,truong2020glu, truong2021learning}. In this work we propose a Gaussian Process (GP) formulation of the matching problem, as detailed in Section~\ref{sec:global_matcher}.

\parsection{Match Refinement}
For detector-free methods, match refinement is typically performed by extracting patches around the sparse matches. Zhou~\etal~\cite{zhou2021patch2pix} propose to refine matches by CNN regression. Sun~\etal~\cite{sun2021loftr} use transformers, with additional improvements by later work~\cite{tang2022quadtree,wang2022matchformer,chen2022aspanformer}. Dense methods in contrast refine matches by dense warp refinement. Troung~\etal~\cite{truong2020glu,truong2021learning} proposed a local-correlation based warp refinement network. In this work, we propose to use stacked feature maps combined with large depth-wise convolution kernels. Our approach to refinement is described in Section~\ref{sec:refiner}.

\parsection{Match Certainty and Sampling}
Although the dense paradigm provides subpixel-level feature matching capabilities, it also comes with inaccurate correspondences in unmatchable regions, resulting in a need for certainty estimation.
Wiles~\etal~\cite{wiles2021co} proposed an MLP-based regressor to infer the matchability potential of dense feature descriptors.
A matchability branch was employed in DGC-Net~\cite{melekhov2019dgc} aiming at predicting the presence or the absence of a pixel correspondence between the images in the form of a binary mask. Recently, in PDC-Net\cite{truong2021learning} and its extension PDC-Net+\cite{truong2021pdc}, the warp estimation was formulated in a probabilistic manner, thus pairing the proposed feature correspondences along with certainty estimates by means of mixture models. We found, however, that their estimated certainty is often confident for unmatchable pairs (Figure~\ref{fig:certainty}). In this work, we propose to model certainty as the likelihood of a pixel having a consistent pairwise match in terms of 3D reconstruction, which provides potent certainty maps as illustrated in Figure~\ref{fig:qualitative}. 
However, in downstream tasks,~\eg, relative pose, the reliability of the extracted correspondence is not the sole factor influencing the performance. For uncalibrated estimation, planar warps are a well known degenerate case~\cite{Chum2005}, and even in the calibrated case the five-point problem is often ill-conditioned~\cite{cavalli2022nefsac,fan2022instability}. Hence, well distributed matches are important for estimation~\cite{hedborg2009fast,barath2022learning}. Motivated by this, we propose a balanced sampling mechanism that provides the estimator with diverse matches. We describe the certainty estimation and balanced sampling in more detail in Section~\ref{sec:sampling}.

\section{Method}
\label{sec:Method}
\begin{figure*}
    \centering
    \includegraphics[width=\linewidth]{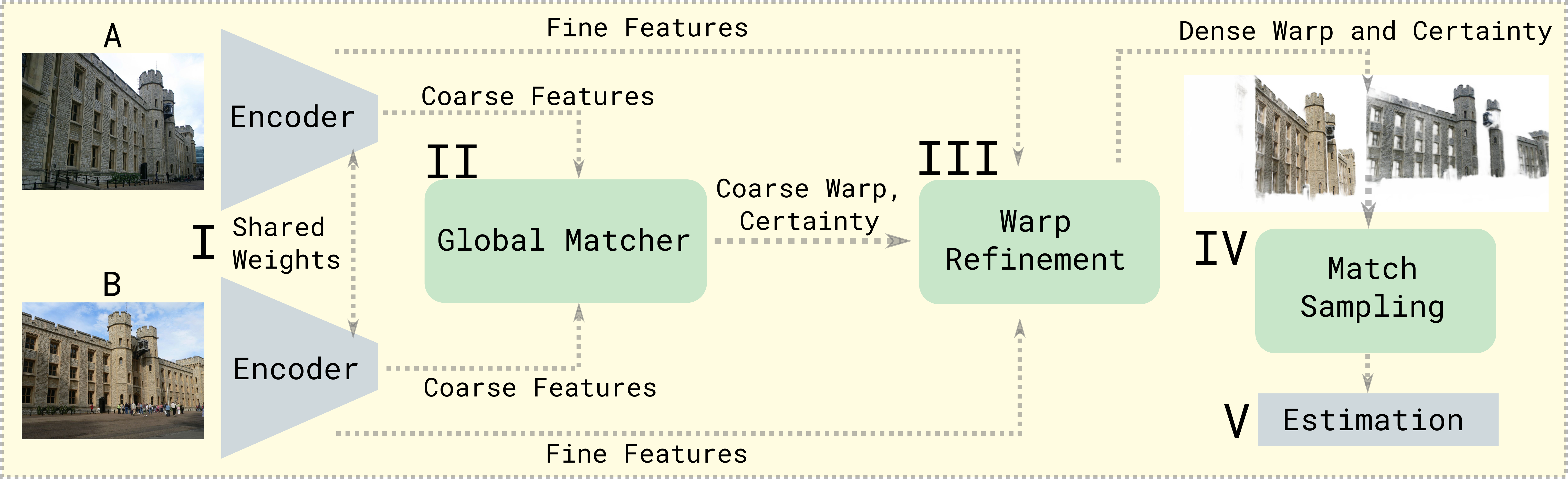}
    \caption{An overview of geometry estimation by dense matching.
    \textbf{I:} In the first stage, a multistride feature pyramid is extracted. We follow previous approaches and use ResNet encoders with shared weights.
    \textbf{II:} In the second stage coarse global matches are established. We improve this stage by viewing it as a embedded probabilistic regression problem combined with a strong embedding decoder. We describe our approach in more detail in Section~\ref{sec:global_matcher}.
    \textbf{III:} The coarse warp is then refined. We propose a stacked feature map approach combined with large depthwise kernels, which increases performance. This is detailed in Section~\ref{sec:refiner}.
    \textbf{IV:} Finally, for geometry estimation a robust certainty estimate is crucial for selecting a set of reliable matches. We find that letting the network learn to classify consistent depth yields a trustworthy certainty estimate. Further combining this with balanced sampling yields even better results. We discuss this in Section~\ref{sec:sampling}.
    \textbf{V:} Once a set of matches have been selected, we use standard robust solvers for estimation as previous methods.}
    \label{fig:method}
\end{figure*}
In the following sections we describe our approach to geometry estimation by dense matching. For an overview, see Figure~\ref{fig:method}. We first provide a general overview of the dense matching framework (Section~\ref{sec:preliminaries}). We then describe our approach for improving the global matcher $G_{\theta}$ (Section~\ref{sec:global_matcher}), the warp refiners $R_{\theta}$ (Section~\ref{sec:refiner}), and certainty estimation along with match sampling~(Section~\ref{sec:sampling}). Lastly, we discuss our loss formulation~(Section~\ref{sec:loss_formulation}).

\subsection{Preliminaries}
\label{sec:preliminaries}
In this paper we consider the task of estimating 3D scene geometry from two images $({I}^{\A},{I}^{\B})$. For matching we choose the dense feature matching paradigm,~\ie, to estimate a dense warp $\text{W}^{\A\to\B}$ and a dense certainty $p^{\A\to\B}$, that is zero for unmatchable pixels.
From this complete set of certain and uncertain matches, a subset of matches are sampled (without replacement). Finally, a robust estimation method is used to infer the geometry from the sampled matches. The task can be divided into five stages. 

In stage \textbf{I}, a feature pyramid is extracted for~\A~and \B, 
\begin{equation}
 \{{\varphi}^{\A}_l\}_{l=1}^L = F_{\theta}({I}^{\A}) \enspace,\enspace \{{\varphi}^{\B}_l\}_{l=1}^L = F_{\theta}({I}^{\B})\enspace,
\end{equation}
where $F_{\theta}$ is an encoder (we use a ResNet50~\cite{he2016deep} pretrained on ImageNet-1K~\cite{russakovsky2015imagenet}), and $l\in \{1,\ldots,L\}$ are the indices for the multiscale features (in our approach $l=1$ corresponds to the rgb values of stride 1, and $l=L$ corresponds to deep features  of stride $2^{L-1} = 32$). We denote the coarse features as $(\varphi_{\text{coarse}}^{\A},\varphi_{\text{coarse}}^{\B})$ and fine features as $(\varphi_{\text{fine}}^{\A},\varphi_{\text{fine}}^{\B})$. In this work the coarse features correspond to stride $\{32, 16\}$ and the fine features to $\{8, 4, 2, 1\}$.

In stage \textbf{II}, we estimate a coarse global warp and certainty from the deep features with a global matcher $G_{\theta}$. Here potential global matches are embedded by the embedder $E_{\theta}$.  We propose to construct the embeddings as solutions to a probabilistic regression problem using a Gaussian Process (GP) formulation. After the embeddings have been computed, an embedding decoder $D_{\theta}$ decodes the embeddings into a dense warp and certainty,~\ie,
\begin{equation}
\left\{
    \begin{aligned}
    \big(\hat{\text{W}}^{\A\to\B}_{\text{coarse}},\enspace \hat{p}^{\A\to\B}_{\text{coarse}}\big) &= G_{\theta}(\varphi^{\A}_{\text{coarse}},\varphi^{\B}_{\text{coarse}}),\enspace\\
    G_{\theta}({\varphi}^{\A}_{\text{coarse}},{\varphi}^{\B}_{\text{coarse}}) &= D_{\theta}\big(E_{\theta}({\varphi}^{\A}_{\text{coarse}},{\varphi}^{\B}_{\text{coarse}})\big).
\end{aligned}
\right.
\end{equation}
We describe our approach to global matching in detail in Section~\ref{sec:global_matcher}.

In stage \textbf{III}, we refine the coarse warp of $G_{\theta}$,~\ie,
\begin{equation}
    \big(\hat{\text{W}}^{\A\to\B},\, \hat{p}^{\A\to\B}\big) = R_{\theta}\big(\varphi_{\text{fine}}^{\A},\varphi_{\text{fine}}^{\B},\hat{\text{W}}^{\A\to\B}_{\text{coarse}},\hat{p}^{\A\to\B}_{\text{coarse}}\big)\,,
\end{equation}
where $\hat{\text{W}}$ is the predicted warp, $\hat{p}$ is the predicted certainty, and $R_{\theta}$ is a set of refiners. This is typically done by local correlation volume refinement. In this work we additionally stack the warped feature maps of \B, and use large depthwise convolution kernels. We describe our approach in detail in Section~\ref{sec:refiner}.
\begin{figure*}
    \centering
    \includegraphics[width=\linewidth]{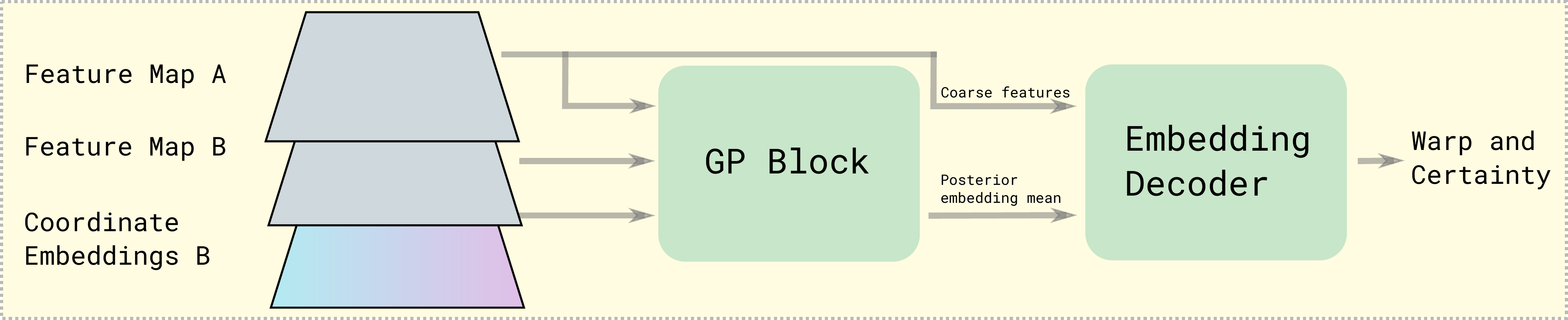}
    \caption{Illustration of the proposed global matcher. The GP, given features and coordinate embeddings, produces a predicive posterior for the warp. The embedding decoder then finds the most likely warp and certainty over the grid in image \A. This is done both at stride 32 and 16. For more details, see Section~\ref{sec:global_matcher}.}
    \label{fig:global_matcher}
\end{figure*}

In stage \textbf{IV}, reliable and accurate matches need to be selected for estimation of scene geometry. For sparse methods this is done at the coarse level by mutual nearest neighbour matching and certainty thresholding. For dense matching, we are free to choose any method, which is an advantage. In this work we do this by sampling the estimated warp and propose a balanced sampling approach. We describe this in Section~\ref{sec:sampling}.

Finally, in stage \textbf{V}, a robust estimator is used to estimate geometry. We use RANSAC with minimal solvers like previous work.
\subsection{Constructing the Global Matcher $G_{\theta}$}
\label{sec:global_matcher}
For an overview of the proposed global matcher, see Figure~\ref{fig:global_matcher}. 

\parsection{Global Matching as Regression}
In this work we construct the global match embeddings as the solution to a (embedded) coordinate regression problem.
We phrase this problem as finding a mapping $\varphi \to \chi$ where $\chi$ are (embeddings of) spatial coordinates in image \B.
We can choose any suitable regression framework to infer the mapping for the pixels in \A. In this work we consider GP regression.

In GP regression, the output (embedded coordinates) ${\chi} \in \mathbb{R}^{H\cdot W\times C}$ is regarded as a collection of random variables, with the main assumption being that these are jointly Gaussian. A GP is uniquely\footnote{With the common assumption that the mean function is 0.} defined by its kernel
that defines the covariance between outputs, and hence must be a positive-definite function to be admissible. We choose the common assumption~\cite{vecvalgp} that the coordinate embedding dimensions are uncorrelated, which makes the kernel block diagonal.  
We choose the exponential cosine similarity kernel~\cite{liu2021swinv2}, which is defined by
\begin{equation}
    k(\varphi,\varphi') = \exp(-\tau)\exp\bigg(\tau\frac{\langle \varphi,\varphi' \rangle}{\sqrt{\langle \varphi,\varphi \rangle \langle \varphi',\varphi' \rangle+\varepsilon}}\bigg),
\end{equation}
since we empirically found it to work well. We found the squared exponential kernel to perform similarly in early experiments, and other kernels could also be considered. We initialize $\tau=5$ and keep it fixed and set $\varepsilon=10^{-6}$. We found that letting the kernel temperature $\tau$ be learnable had negligible effect on the performance, and that our method was robust to initializations for $\tau \in [3,10]$.

With the standard assumption~\cite{rasmussen2003gaussian} that the measurements $({\varphi}_{\text{coarse}}^{\B},{\chi}_{\text{coarse}}^{\B})$ are observed with i.i.d.\ noise, the analytic formulae for the posterior conditioned on the features of \B~are given by
\begin{equation}
\left\{
\begin{aligned}
    \mu({\varphi}^{\A}_{\text{coarse}}|{\varphi}^{\B}_{\text{coarse}}) &= K^{\A\B}(K^{\B\B}+\sigma_n^2 I)^{-1}{\chi}^{\B}_{\text{coarse}},\\
    \Sigma({\varphi}^{\A}_{\text{coarse}}|{\varphi}^{\B}_{\text{coarse}}) &= K^{\A\A}-K^{\A\B}(K^{\B\B}+\sigma_n^2 I)^{-1}K^{\B\A},
\end{aligned}
\right.
\end{equation}
where $K$ denotes the kernel matrix, $\mu$ is the posterior mean function, $\sigma_n = 0.1$ is the standard deviation of the measurement noise, and $\Sigma$ is the posterior covariance. We refer to Rasmussen~\cite{rasmussen2003gaussian} for details on GP regression. 

\begin{figure*}
    \centering
    \includegraphics[width=\linewidth]{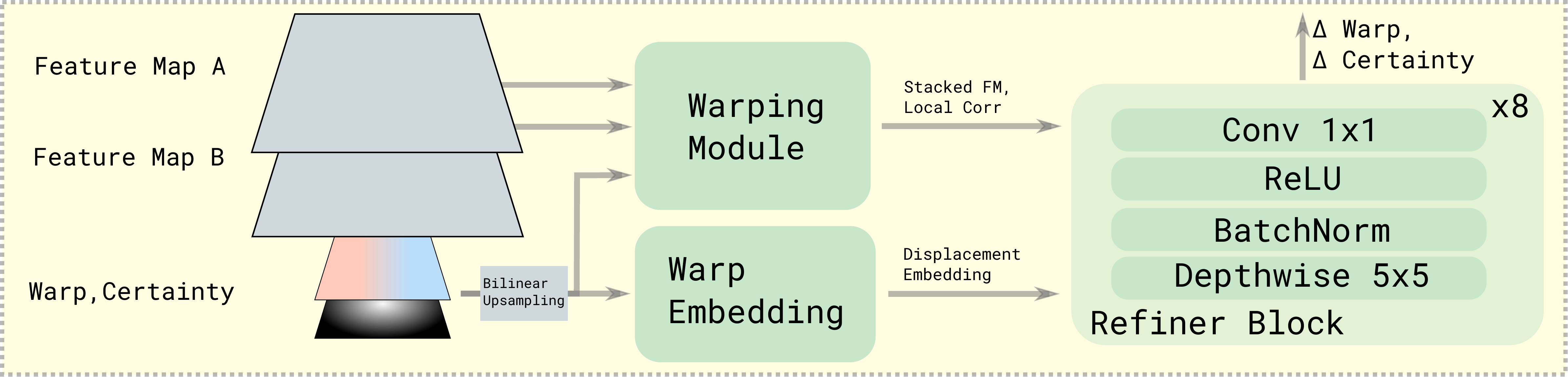}
    \caption{Illustration of the proposed Warp Refiners. Warp Refiners take in fine features, and the upsampled coarse warps and certainty estimates. The coarse warp is used both to warp the \B \ features directly to the \A \  feature grid, as well as being used to construct a local correlation volume around the warp target in the image \B. Furthermore the warp itself is converted to a displacement, and linearly embedded. These features combined are concatenated and fed into the refiner blocks. For more details, see Section~\ref{sec:refiner}.}
    \label{fig:refiner}
\end{figure*}

\parsection{Coordinate Embeddings}
One issue with coordinate regression is how to deal with multimodality. GP posteriors are unimodal in the output space, and hence multimodal matches can degrade performance. 

To deal with this issue we use a cosine embedding
\begin{equation}
    {B}_{\mathcal{F}}(x;W,b) = \cos(Wx+b),
\end{equation}
where $x\in \mathbb{R}^2$ is the image coordinate, $W_{ij}\sim \mathcal{N}(0,\ell^2)$, $b_i\sim\mathcal{U}_{[0,2\pi]}$, $i\in\{1,\hdots,C\}$, $j\in\{1,2\}$.
These types of embeddings are well known to preserve multimodality~\cite{snippe1992discrimination}, and possess multiple other nice properties~\cite{rahimi2007random,tancik2020fourier}.
\begin{figure}
    \centering
    \includegraphics[width=\linewidth]{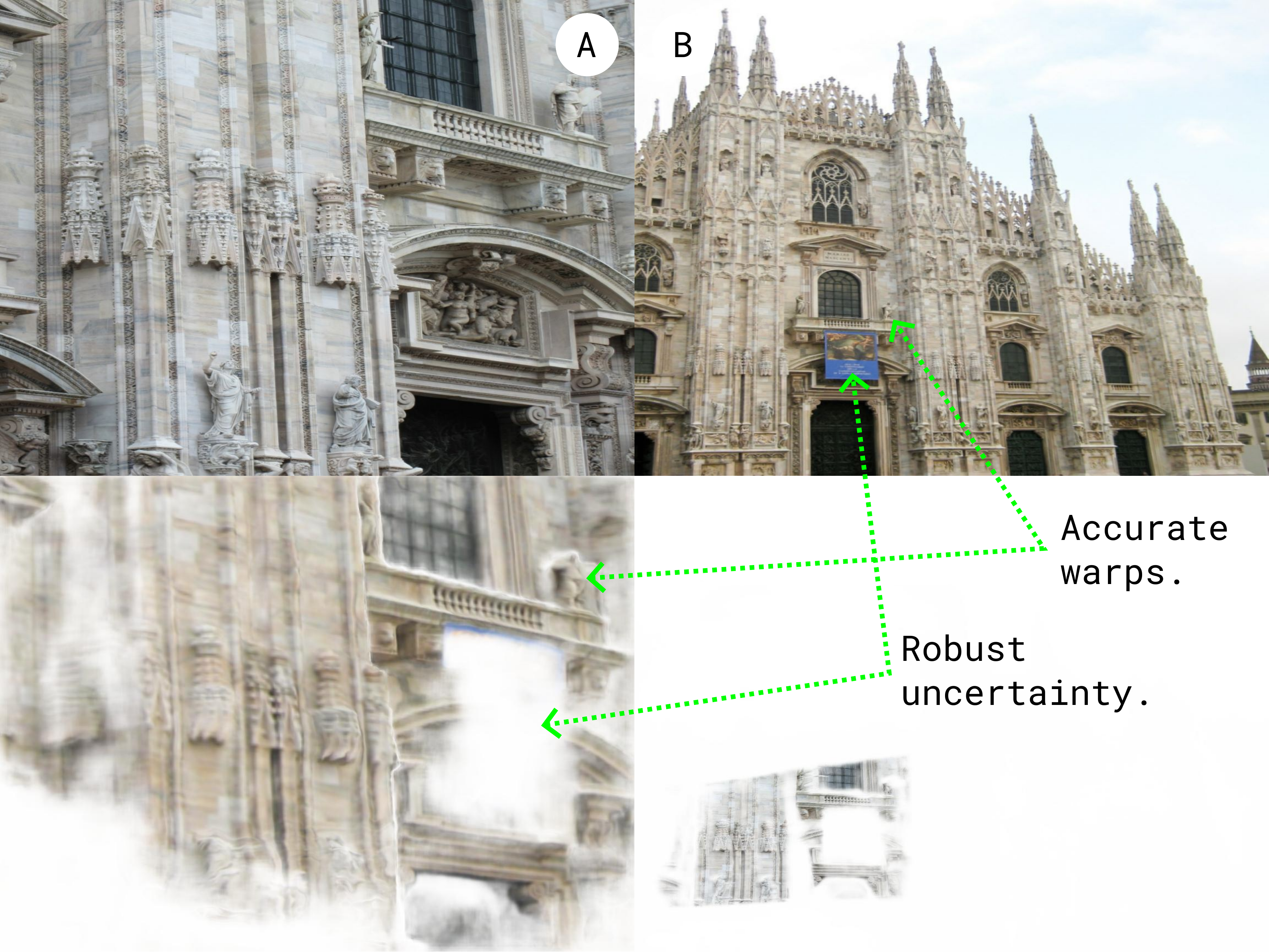}
    \caption{Dense methods often struggle with large viewpoint changes. Our proposed global matcher + refiner architecture is able to produce accurate warps and certainty even for extreme perspective. Top row, image \A~and \B. Bottom row, forward and reverse warp weighted by certainty.}
    \label{fig:extreme_viewpoint}
\end{figure}

\parsection{Embedding Decoder}
While the embedded regression yields a powerful probabilistic representation of the warp, most dense methods require a unimodal warp estimate for the subsequent refinement steps. There are multiple ways of decoding coordinates from the posterior. We use a simple method of reshaping the predictive mean back into grid form $\mu_{\text{grid}}({\varphi}^{\A}_{\text{coarse}}|{\varphi}^{\B}_{\text{coarse}})\in \mathbb{R}^{H_{\text{coarse}}\times W_{\text{coarse}} \times C}$ and let
\begin{equation}
    G_{\theta}({\varphi}^{\A}_{\text{coarse}},{\varphi}^{\B}_{\text{coarse}}) =  D_{\theta}(\mu_{\text{grid}}({\varphi}^{\A}_{\text{coarse}}|{\varphi}^{\B}_{\text{coarse}}), {\varphi}^{\A}_{\text{coarse}}),
\end{equation} where $D_{\theta}$ is an embedding decoder. The decoder predicts coordinates in the canonical grid $[-1,1]\times[-1,1]$, and additionally logits for the predicted validity of the matches, for each pixel. The architecture of the embedding decoder is inspired by the decoder proposed by Yu~\etal~\cite{yu2018learning}. We use global matchers on both stride 32 and 16 features of the backbone, and the stride 16 embedding decoder takes in context feature maps from the stride 32 decoder.

\subsection{Refining the Warp with $R_{\theta}$}

\label{sec:refiner}
Once the embeddings have been decoded, we refine the warp using CNN refiners similarly to previous work~\cite{truong2020glu,shen2020ransac}. They take as input the feature maps and the previous warp and certainty. The warp and certainty are bilinearly upsampled to match the scale of the feature maps. These predict a residual offset for the estimated warp, and a logit offset for the certainty. The process is repeated until we reach full resolution. The process is described recursively by
\begin{equation}
    \big(\hat{W}^{\A\to\B}_l,\;\hat{p}_{l}^{\A\to\B}\big)   = R_{\theta,l}(\varphi_l^{\A},\varphi_l^{\B},\hat{W}^{\A\to\B}_{l+1},\hat{p}_{l+1}^{\A\to\B}).
\end{equation}
Compared to previous work, we make improvements to both the input representations and the architecture of the refiners.
Previous work~\cite{truong2021learning,truong2021pdc} uses the warp, the feature maps of \A, and local correlation in \A~with warped feature maps from \B. In contrast, we use all channels of the warped feature maps of \B~by simple concatenation, as well as local correlation in \B~instead of \A. We investigate the effect of this change of representation in Table~\ref{tab:abl_refiners} and find that it yields improvements in warp accuracy.

Finally, we improve the architecture of the refiner blocks themselves. Previous work~\cite{truong2020glu,truong2021learning} uses a DenseNet~\cite{huang2017densely} architecture with 3x3 non-separable kernels. We instead propose to use bigger 5x5 depthwise separable kernels, followed by a 1x1 convolution. As we show in Table~\ref{tab:abl_refiners}, this improvement leads to large gains in performance. Empirically we found 8 refiner blocks per scale to give the best performance. The architecture is detailed in Figure~\ref{fig:refiner}. We qualitatively show the high robustness and accuracy of DKM warps in Figure~\ref{fig:extreme_viewpoint}.

\subsection{Certainty Estimation and Sampling for Geometry Estimation}
\label{sec:sampling}
\begin{figure}
    \centering
    \includegraphics[width=\linewidth]{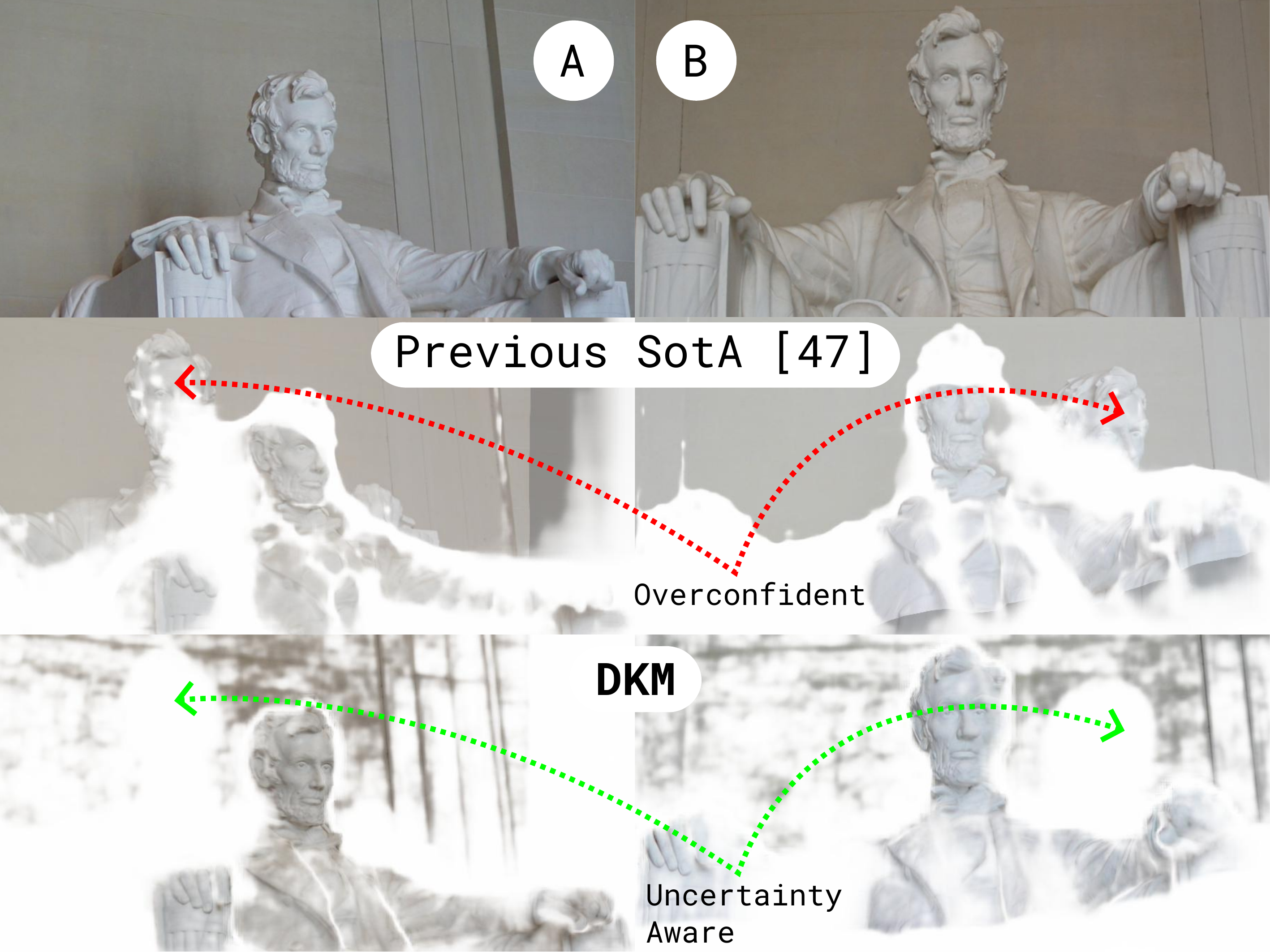}
    \caption{Qualitative Comparison of our certainty estimate compared to PDC-Net+. Top row, image \A, image \B. Middle row, results for PDC-Net+. Bottom row, results for DKM. DKM places high certainty on repeatable matches, while PDC-Net+ is often overconfident in untextured regions, even predicting high certainty for non-matchable pixel-pairs.}
    \label{fig:certainty}
\end{figure}

\parsection{Certainty Estimation by Classifying Depth-consistent Matches} We leverage the rich 3D models and densified depth maps in the large scale MegaDepth~\cite{li2018megadepth} dataset. We find consistent matches first by warping $\A\to\B$ using the ground truth depth, and then applying a relative depth consistency constraint in image $\B$. This equates to
\begin{equation}
    p^{\A\to\B} = \bigg|\frac{z^{\A\to\B}-z^{\B}}{z^{\B}}\bigg| <\alpha
\end{equation}
where $z$ is the depth, $z^{\A\to\B}$ depth projected using the ground truth 3D model, and $\alpha=0.05$.
This approach has similarities to the approach in LoFTR~\cite{sun2021loftr}, but they instead indirectly apply the constraint by finding mutual nearest neighbours.
We demonstrate the importance of a good certainty estimate in Table~\ref{tab:abl_sampling}, and show a qualitative comparison of our certainty estimate compared to the previous best perfoming dense work PDC-Net+~\cite{truong2021pdc} in Figure~\ref{fig:certainty}.

\parsection{Sampling Balanced Matches}
For estimation, match sampling is required. A simple approach is to sample using the estimated warp certainty as weight. This approach is written as,
\begin{equation}
\label{eq:sample_matches}
    \{x^{\A}_i,x^{\B}_i\}_{i=1}^N \sim \hat{p}^{\A\to\B}.
\end{equation}

Like previous semi-sparse~\cite{sun2021loftr,chen2022aspanformer} and dense works~\cite{truong2021pdc} we threshold the estimated certainty. We use a threshold of $0.05$, and sample matches from the thresholded distribution. 

While certainty weighted sampling produces good matches, having diverse matches typically improves estimation~\cite{Chum2005,hedborg2009fast,fan2022instability,cavalli2022nefsac}. To achieve this, we propose a simple method for producing scene balanced matches. First, we sample a large set of matches using the estimated certainty. Secondly, we compute a kernel density estimate (KDE) in the 4-dimensional match space. Thirdly, we weight each match with the reciprocal of the KDE to produce a balanced set of samples. This procedure produces a balanced distribution in the scene. We investigate the impact of the balanced sampling in Table~\ref{tab:abl_sampling}, and find that it improves performance.

\subsection{Loss Formulation}
\label{sec:loss_formulation}
Like previous work~\cite{shen2020ransac,sarlin2020superglue,truong2021learning} we use separate losses for each stride $l\in \{1,...,L\}$, and use a combination of regression and certainty~\cite{melekhov2019dgc,zhou2021patch2pix,tan2022eco} losses to train our model. The combined loss is
\begin{equation}
    \mathcal{L} = \sum_{l=1}^L\mathcal{L}_{\text{warp}}(\hat{W}_l^{\A\to\B}) + \lambda \mathcal{L}_{\text{conf}}(\hat{p}_l^{\A\to\B}),
\end{equation}
where $\lambda = 0.01$ is a balancing term, similarly to~\cite{melekhov2019dgc,tan2022eco}.

Specifically, for the warp loss we use the $\ell_2$ distance between the predicted and ground truth warp, as in~\cite{sun2021loftr}. For the certainty loss we use the unweighted binary cross entropy between the predicted certainty and the ground truth consistent depth mask.
Our losses at a given stride $l$ are
\begin{align}
\label{eq:warploss}
    \mathcal{L}_{\text{warp}}(\hat{W}_l^{\A\to\B}) &= \sum_{\text{grid}}{p_l\odot \norm[\big]{W_l^{\A\to\B}-\hat{W}_l^{\A\to\B}}_2},\\
    \mathcal{L}_{\text{conf}}(\hat{p}_l) &= \sum_{\text{grid}} p_l\log{\hat{p}_l} + (1-p_l)\log{(1-\hat{p}_l)},
\end{align}
where the summation is done over the image grid in \A. Like Zhou~\etal~\cite{zhou2021patch2pix} we set $p$ in the fine stride loss to 0 whenever the coarse stride warp is outside a threshold distance from the ground truth. We further found it beneficial to detach the gradients between scales.

\section{State-of-the-Art Comparison}
\label{sec:sota}
Similarly to previous approaches~\cite{sarlin2020superglue,sun2021loftr,tang2022quadtree,chen2022aspanformer}, we train and evaluate our approach separately on \textbf{outdoor} and \textbf{indoor} geometry estimation. For evaluation we present the average of 5 benchmark runs. For DKM we sample a maximum of 5000 matches. 

\subsection{Training Details}
\label{sec:training}
We use a batch size of 32 with a learning rate of $4\cdot10^{-4}$ for the decoder and refiners, and $2\cdot10^{-5}$ for the backbone. We use the AdamW~\cite{loshchilov2018decoupled} optimizer with a weight-decay factor of $10^{-2}$. We train for \num{250000} steps, decaying the learning rate by a factor $0.2$ at step \num{166666} and \num{225000}.
Training takes roughly 5 days on 4 A100fat GPUs, which is comparable to LoFTR that converges in 1 day on 64 1080ti GPUs.

\parsection{Outdoor Training}
We train on the real world dataset MegaDepth~\cite{li2018megadepth}, using the same training and test split as in previous work~\cite{sun2021loftr,chen2022aspanformer}. We resize the images to a fixed resolution of $540\times 720$.

\parsection{Indoor Training}
For indoor two-view pose estimation we additionally train on the ScanNet~\cite{dai2017scannet} dataset in a similar fashion as previous work~\cite{sarlin2020superglue,sun2021loftr} and use a resolution of $480 \times 640$.

\subsection{Outdoor Geometry Estimation}
\parsection{HPatches Homography}
\begin{table}
    \small
    \centering
    \caption{Homography estimation on HPatches, measured in AUC (higher is better). The top portion contains sparse methods, while the bottom portion contains dense methods}
    \begin{tabular}{llll}
    \toprule
     Method $\downarrow$\quad\quad\quad AUC $\rightarrow$       & $@3$px & $@5$px & $@10$px\\
       \midrule
       SuperGlue~\cite{sarlin2020superglue}~\tiny{CVPR'19} & 53.9 & 68.3 & 81.7 \\
LoFTR~\cite{sun2021loftr}~\tiny{CVPR'21} & 65.9 & 75.6 & 84.6 \\
TopicFM~\cite{giang2022topicfm}~\tiny{Arxiv'22} & 67.3 & 77.0 & 85.7 \\
    3DG-STFM~\cite{mao20223dg}~\tiny{ECCV'22} & 64.7 & 73.1 & 81.0 \\
    ASpanFormer~\cite{chen2022aspanformer}~\tiny{ECCV'22}  & 67.4 & 76.9 & 85.6\\

    \midrule
    PDC-Net+~\cite{truong2021pdc}~\tiny{Arxiv'21} & 67.7 & 77.6 & 86.3 \\
\textbf{DKM} & \textbf{71.3} & \textbf{80.6} & \textbf{88.5} \\
    \bottomrule
    \end{tabular}
    \label{tab:hpatches_homog}
\end{table}
HPatches~\cite{balntas2017hpatches} depicts planar scenes divided in sequences, with transformations restricted to homographies. We follow the evaluation protocol proposed LoFTR~\cite{sun2021loftr}, resizing the shorter side of the images to 480. Table~\ref{tab:hpatches_homog} clearly shows the superiority of DKM, showing gains of +3.6$\text{ AUC}@\text{3px}$ compared to the best previous method.

\begin{table}
\small
    \centering
    \caption{ Pose estimation results on the Megadepth-1500 benchmark, measured in AUC (higher is better). The top portion contains sparse methods, while the bottom portion contains dense methods.}
    \begin{tabular}{l lll}
    \toprule
     Method $\downarrow$\quad\quad\quad AUC $\rightarrow$& $@5^{\circ}$&$@10^{\circ}$&$@20^{\circ}$\\
     \midrule
         SuperGlue~\cite{sarlin2020superglue}~\tiny{CVPR'19} & 42.2 & 61.2 & 76.0 \\
         LoFTR~\cite{sun2021loftr}~\tiny{CVPR'21}& 52.8 & 69.2 & 81.2 \\
         QuadTree~\cite{tang2022quadtree}~\tiny{ICLR'22} & 54.6	& 70.5	&82.2\\
         MatchFormer~\cite{wang2022matchformer}~\tiny{ACCV'22} & 52.9  & 69.7 & 82.0\\
         TopicFM~\cite{giang2022topicfm}~\tiny{Arxiv'22} & 54.1 & 70.1 & 81.6 \\
        3DG-STFM~\cite{mao20223dg}~\tiny{ECCV'22} & 52.6 & 68.5 & 80.0 \\
        
         ASpanFormer~\cite{chen2022aspanformer}~\tiny{ECCV'22} & 55.3  & 71.5 & 83.1\\
        \midrule
        PDC-Net+~\cite{truong2021pdc}~\tiny{Arxiv'21} & 51.5 & 67.2 & 78.5 \\
        DenseGAP~\cite{kuang2021densegap}~\tiny{ICPR'22} & 41.2 & 56.9 & 70.2\\
        ECO-TR~\cite{tan2022eco}~\tiny{ECCV'22} &
48.3 & 65.8 & 78.5\\
         \textbf{DKM}  &  \textbf{60.4} & \textbf{74.9} & \textbf{85.1}\\

    \bottomrule

    \end{tabular}

    \label{tab:megadepth-loftr}
\end{table}
\parsection{MegaDepth-1500 Pose Estimation}
We use the MegaDepth-1500 test set~\cite{sun2021loftr} which consists of 1500 pairs from scene 0015 (St.\ Peter's Basilica) and 0022 (Brandenburger Tor). We follow the protocol in~\cite{sun2021loftr,chen2022aspanformer} and use a RANSAC threshold of 0.5 with intrinsics equivalent to a longer side of 1200. Our results, presented in Table~\ref{tab:megadepth-loftr}, show that our method sets a new state-of-the-art. Notably, we outperform the current best sparse method ASpanFormer~\cite{wang2022matchformer} with an improvement of +4.9 AUC$@5^{\circ}$. Furthermore, we significantly outperform the best previous dense method PDC-Net+~\cite{truong2021pdc} with an impressive improvement of +8.9 AUC$@5^{\circ}$.

\parsection{Additional Benchmarks}
We create a novel benchmark based on 8 diverse MegaDepth scenes, where we show major improvements. We further do additional comparisons to COTR~\cite{jiang2021cotr} and ECO-TR~\cite{tan2022eco} on the St.\ Paul's Cathedral scene, with DKM showing large improvements. The details of both these experiments can be found in supplementary material~\ref{sec:mega-8-scenes} and~\ref{sec:st-pauls} respectively.

\subsection{Indoor Geometry Estimation}

\parsection{ScanNet-1500 Pose Estimation} ScanNet~\cite{dai2017scannet} is a large scale indoor dataset, composed of challenging sequences with low texture regions and large changes in perspective. We follow the evaluation in SuperGlue~\cite{sarlin2020superglue}. Results are presented in Table~\ref{tab:ScanNet}. Our model achieves a +4.0 AUC$@5^{\circ}$ gain compared to the previous best sparse method. Compared to the previous best dense method our performance gains are even larger, with gains of +9.3. 
\begin{table}[h!]
\small
    \centering
    \caption{ Pose estimation results on the ScanNet-1500 benchmark, measured in AUC (higher is better). The upper portion contains sparse and semi-sparse methods, while the lower portion contains dense methods.}
    \begin{tabular}{l lll}
    \toprule
     Method $\downarrow$\quad\quad\quad AUC $\rightarrow$& $@5^{\circ}$&$@10^{\circ}$&$@20^{\circ}$\\
     \midrule
SuperGlue~\cite{sarlin2020superglue}~\tiny{CVPR'19}  & 16.2&	33.8&	51.8\\
          LoFTR~\cite{sun2021loftr}~\tiny{CVPR'21} & 22.1 &	40.8&	57.6\\
          QuadTree~\cite{tang2022quadtree}~\tiny{ICLR'22} & 24.9	& 44.7	&61.8\\
          MatchFormer~\cite{wang2022matchformer}~\tiny{ACCV'22} &24.3 & 43.9 & 61.4 \\
    3DG-STFM~\cite{mao20223dg}~\tiny{ECCV'22} & 23.6 & 43.6 & 61.2 \\
     ASpanFormer~\cite{chen2022aspanformer}~\tiny{ECCV'22} & 25.6 & 46.0 & 63.3\\
    \midrule
         PDC-Net~\cite{truong2021learning}~\tiny{CVPR'21} &  18.7 & 37.0 & 54.0 \\
          PDC-Net+~\cite{truong2021pdc}~\tiny{Arxiv'21}& 20.3 & 39.4 & 57.1 \\
    DenseGAP~\cite{kuang2021densegap}~\tiny{ICPR'22} & 16.9 & 34.9 & 53.2  \\
\textbf{DKM}  &  \textbf{29.4} & \textbf{50.7} & \textbf{68.3}\\

    \bottomrule

    \end{tabular}

    \label{tab:ScanNet}
\end{table}

\section{Ablation Study}
\label{sec:ablation}
Next, we investigate design choices of our approach. 

\parsection{Global Matcher}
Here we investigate the performance impact of replacing a strong baseline correlation volume regressor, similar to the one used in~\cite{truong2021learning} with our proposed kernelized regression and embedding decoder approach. The results are shown in Table~\ref{tab:abl_global_matcher}. We see that our proposed method yields an improvement of +1.1 $\text{AUC}@5^{\circ}$, highlighting the benefits of our proposed global matcher. As expected, the linear regression approach instead of cosine embedded coordinates does not perform as well.
\begin{table}[h!]
    \centering
    \caption{Impact of our proposed Global Matcher (GM), using either linear or cosine coordinate embeddings, compared to a strong baseline. Measured in AUC (higher is better).}
    \begin{tabular}{llll}
    \toprule
     GM $\downarrow$\quad\qquad AUC $\rightarrow$& $@5^{\circ}$&$@10^{\circ}$&$@20^{\circ}$\\
       \midrule
    Baseline& 57.0 & 72.1 & 82.9 \\
    Proposed Linear & 57.9 & 72.9 & 83.7 \\
    Proposed Cosine& \textbf{58.1} & \textbf{73.2} & \textbf{83.8} \\
    
    \bottomrule
    \end{tabular}
    \label{tab:abl_global_matcher}
\end{table}

\parsection{Warp Refiners}
Here we ablate both the architecture, and the effect of the features used. For the architecture we exchange the depthwise convolution blocks for refiners used in previous dense matching work~\cite{truong2021learning}. The results of this ablation are shown in Table~\ref{tab:abl_refiners}. Our depthwise refiners significantly outperform the baseline, with a gain of +4.8$\text{ AUC}@5$. Furthermore, we find that our input representation yields an improvement of +1.5$\text{ AUC}@5$.

\begin{table}
    \small
    \centering
    \caption{Impact of our proposed depthwise (DW) warp refiners, and stacked feature map (FM) approach compared to a strong baseline. Measured in AUC (higher is better).}
    \begin{tabular}{llll}
    \toprule
     Warp Refiner $\downarrow$\quad AUC $\rightarrow$& $@5^{\circ}$&$@10^{\circ}$&$@20^{\circ}$\\
       \midrule
    Baseline Refiners & 54.9 & 70.0 & 81.6 \\
    Baseline Inputs & 56.5 & 71.8 & 82.7 \\
    DW Refiners, Stacked FM & \textbf{58.1} & \textbf{73.2} & \textbf{83.8} \\
    \bottomrule
    \end{tabular}
    \label{tab:abl_refiners}
\end{table}

\parsection{Match Sampling}
Here we investigate the impact of the match sampling strategy. First, we compare to a baseline using no certainty estimate. We then ablate the effect of balancing the match sampling using the reciprocal of the KDE estimate. We present results in Table~\ref{tab:abl_sampling}, which clearly shows the need for certainty. We also find that the proposed balanced sampling approach helps in the estimation stage, increasing performance with an improvement of +2.0$\text{ AUC}@5$.

\begin{table}
    \small
    \centering
    \caption{Impact of balanced match sampling for two-view pose estimation, measured in AUC (higher is better).}
    \begin{tabular}{llll}
    \toprule
     Sampling $\downarrow$\quad AUC $\rightarrow$& $@5^{\circ}$&$@10^{\circ}$&$@20^{\circ}$\\
       \midrule
    No Certainty Sampling & 42.9 & 58.1 & 70.4 \\
    Certainty Sampling & 56.1 & 71.7 & 83.0 \\
    Balanced Sampling & \textbf{58.1} & \textbf{73.2} & \textbf{83.8} \\
    \bottomrule
    \end{tabular}
    \label{tab:abl_sampling}
\end{table}

\parsection{Resolution}
Tinchev \etal~\cite{tinchev2020xrcnet} recently noted the importance of increasing input resolution for estimation performance. 
To gauge the effect of resolution on estimation performance in the dense paradigm we trained DKM on a set of different resolutions. We present the results of our study in Table~\ref{tab:abl_resolution}. We find that setting the resolution sufficiently high is important for accurate estimation. In particular, comparing $384\times 512$ to $540\times720$ we find an increase in performance of +1.3 $\text{AUC}@5^{\circ}$. 
\begin{table}
    \small
    \centering
    \caption{Impact of changing training resolution for two-view pose estimation, measured in AUC (higher is better).}
    \begin{tabular}{llll}
    \toprule
     Resolution $\downarrow$ AUC $\rightarrow$& $@5^{\circ}$&$@10^{\circ}$&$@20^{\circ}$\\
       \midrule
    384$\times$512 & 58.1 & 73.2 & 83.8 \\
    480$\times$640 & 58.9 & 73.9 & 84.4 \\
    540$\times$720 & \textbf{59.4} & \textbf{74.0} & \textbf{84.5} \\
    \bottomrule
    \end{tabular}
    \label{tab:abl_resolution}
\end{table}

\parsection{Bidirectionality}
Previous dense work~\cite{truong2021pdc,tan2022eco} has investigated incorporating mutual nearest neighbours in dense matching. Here we propose to instead simply concatenate the reverse warp matches. Results are presented in Table~\ref{tab:abl_bidirectionality}. We find an improvement of +1.0 $\text{AUC}@5^{\circ}$.
\begin{table}
    \small
    \centering
    \caption{Impact of bidirectional DKM for two-view pose estimation, measured in AUC (higher is better).}
    \begin{tabular}{llll}
    \toprule
     Warp $\downarrow$ AUC $\rightarrow$& $@5^{\circ}$&$@10^{\circ}$&$@20^{\circ}$\\
       \midrule
    Unidirectional & 59.4 & 74.0 & 84.5 \\
    Bidirectional & \textbf{60.4} & \textbf{74.9} & \textbf{85.1} \\
    
    \bottomrule
    \end{tabular}
    \label{tab:abl_bidirectionality}
\end{table}

\begin{figure}
    \centering
    \includegraphics[width=\linewidth]{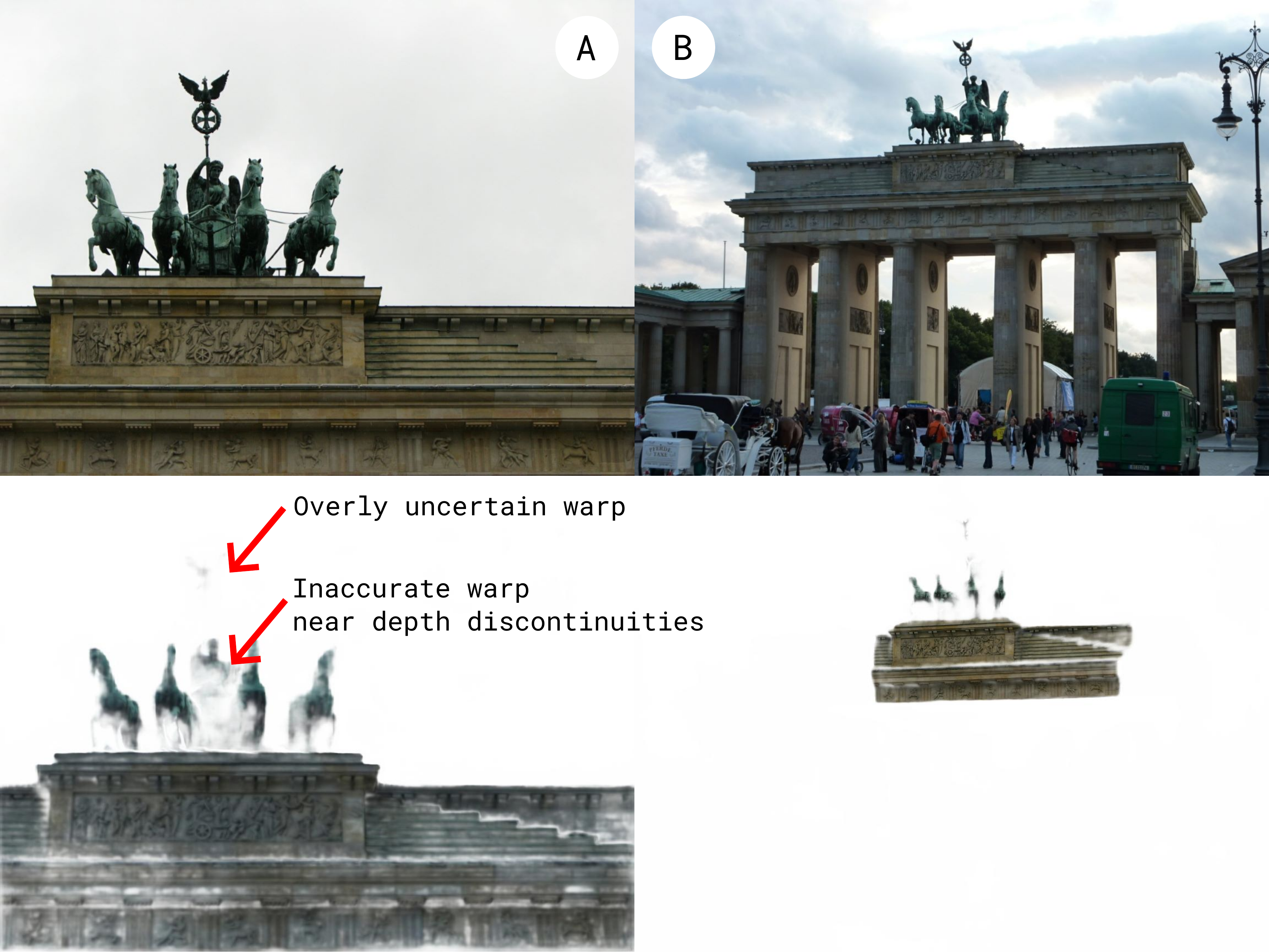}
    \caption{Representative failure case for DKM. Our unimodal warp refinement can struggle near depth-discontinuities, and the proposed certainty estimate is occationally overly uncertain.}
    \label{fig:failure}
\end{figure}
\section{Conclusion}
\label{sec:conclusion}
We have presented \textbf{DKM}, a novel dense feature matching approach that achieves state-of-the-art two-view geometry estimation results. Three distinct contributions were proposed. We proposed a strong global matcher with a kernelized regressor and embedding decoder. Furthermore, we proposed warp refinement through large depth-wise kernels on stacked feature maps. Finally, we proposed a simple way of learning dense confidence maps by directly classifying consistent depth, and a balanced sampling approach for dense warps. Our extensive experiments clearly showed the superiority of our method, with gains of +8.9 $\text{AUC}@5^{\circ}$ on the MegaDepth-1500 benchmark.

\parsection{Limitations}
While our global matcher can gracefully handle multimodality, the proposed dense warp refinement is unimodal. This poses challenges where the warp is discontinuous,~\eg, at depth boundaries. We also found DKM to be overly uncertain for small objects bordering the sky. This could be a limitation of learning to classify consistent depth, instead of predicting model uncertainty as in,~\eg, PDC-Net. We illustrate an example of both these weaknesses in Figure~\ref{fig:failure}.

\section*{Acknowledgements}
This work was partially supported by the Wallenberg Artificial Intelligence,
Autonomous Systems and Software Program (WASP) funded by Knut and Alice Wallenberg Foundation; and by the strategic research environment ELLIIT funded by the Swedish government. The computations were enabled by resources provided by the Swedish National Infrastructure for Computing (SNIC), partially funded by the Swedish Research Council
through grant agreement no.~2018-05973, and by the
Berzelius resource provided by the Knut and Alice Wallenberg
Foundation at the National Supercomputer Centre.

\small{\bibliography{main}
}
\bibliographystyle{plain}
\clearpage
\appendix
\begin{strip}
\vspace{-20pt}
\begin{center}
\textbf{\Large Supplementary Material for}
\end{center}
\begin{center}\textbf{\Large DKM: Dense Kernelized Feature Matching for Geometry Estimation}
\end{center}
\end{strip}

\section{Additional State-of-the-Art Comparison}
\subsection{MegaDepth-8-Scenes Pose Estimation}
\label{sec:mega-8-scenes}
\begin{figure}
    \centering
    \includegraphics[width=\linewidth]{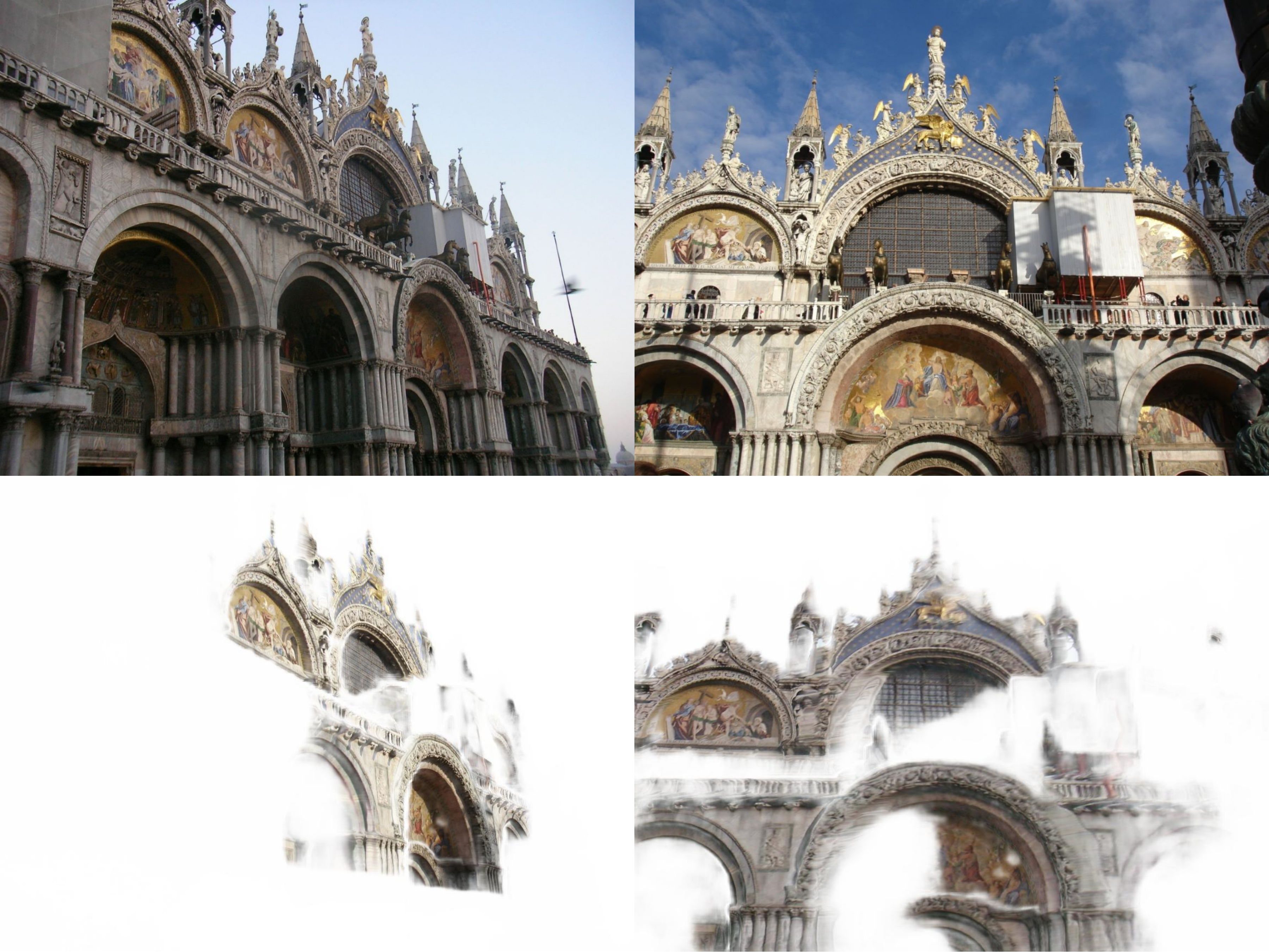}
    \caption{Qualitative example of pair in Piazza San Marco (0008) with DKM warp and certainty.}
    \label{fig:mega0008}
\end{figure}

\begin{figure}
    \centering
    \includegraphics[width=\linewidth]{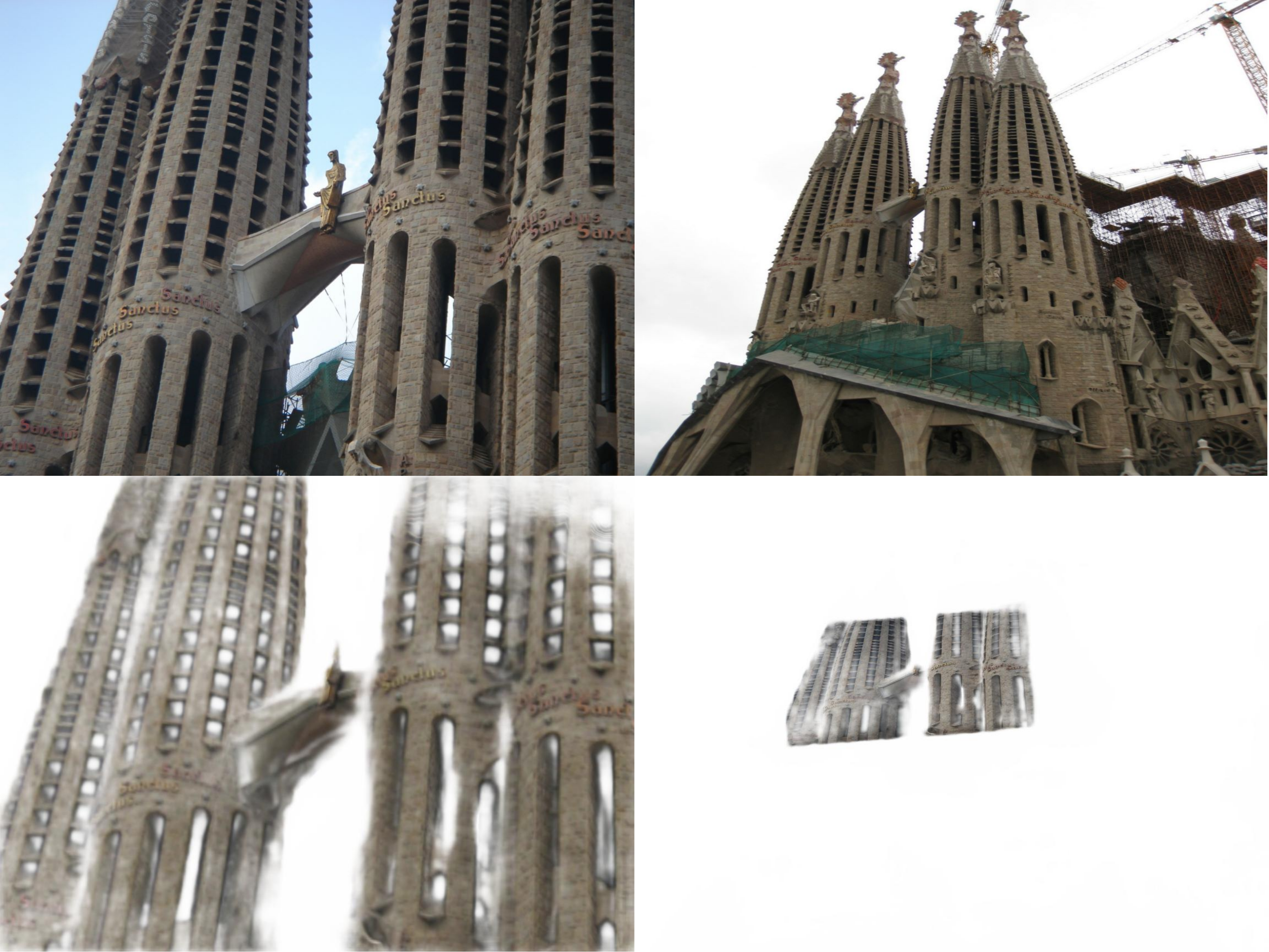}
    \caption{Qualitative example of pair in Sagrada Familia (0019) with DKM warp and certainty.}
    \label{fig:mega0019}
\end{figure}

\begin{figure}
    \centering
    \includegraphics[width=\linewidth]{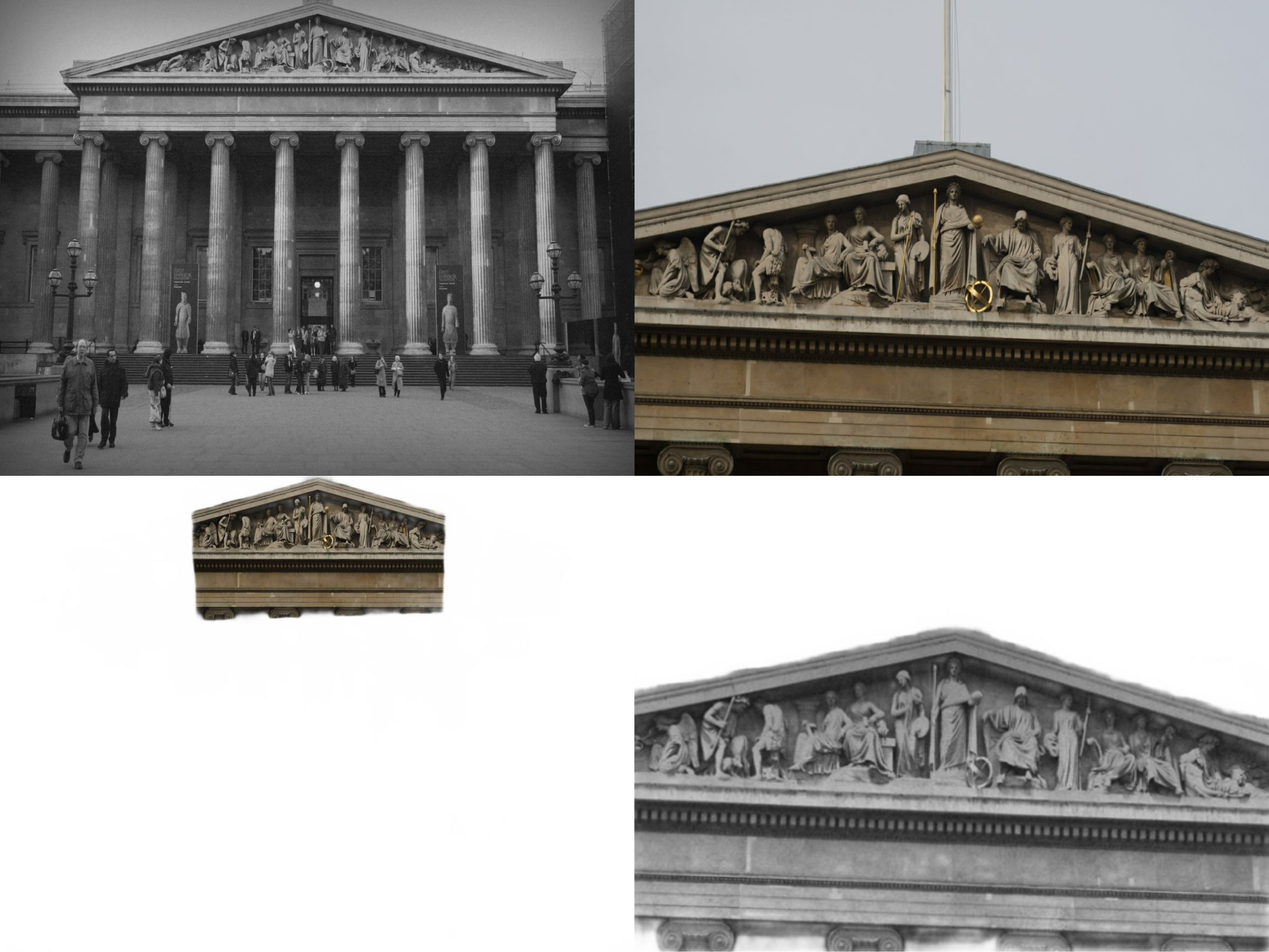}
    \caption{Qualitative example of pair in Britism Museum (0024) with DKM warp and certainty.}
    \label{fig:mega0024}
\end{figure}

\begin{figure}
    \centering
    \includegraphics[width=\linewidth]{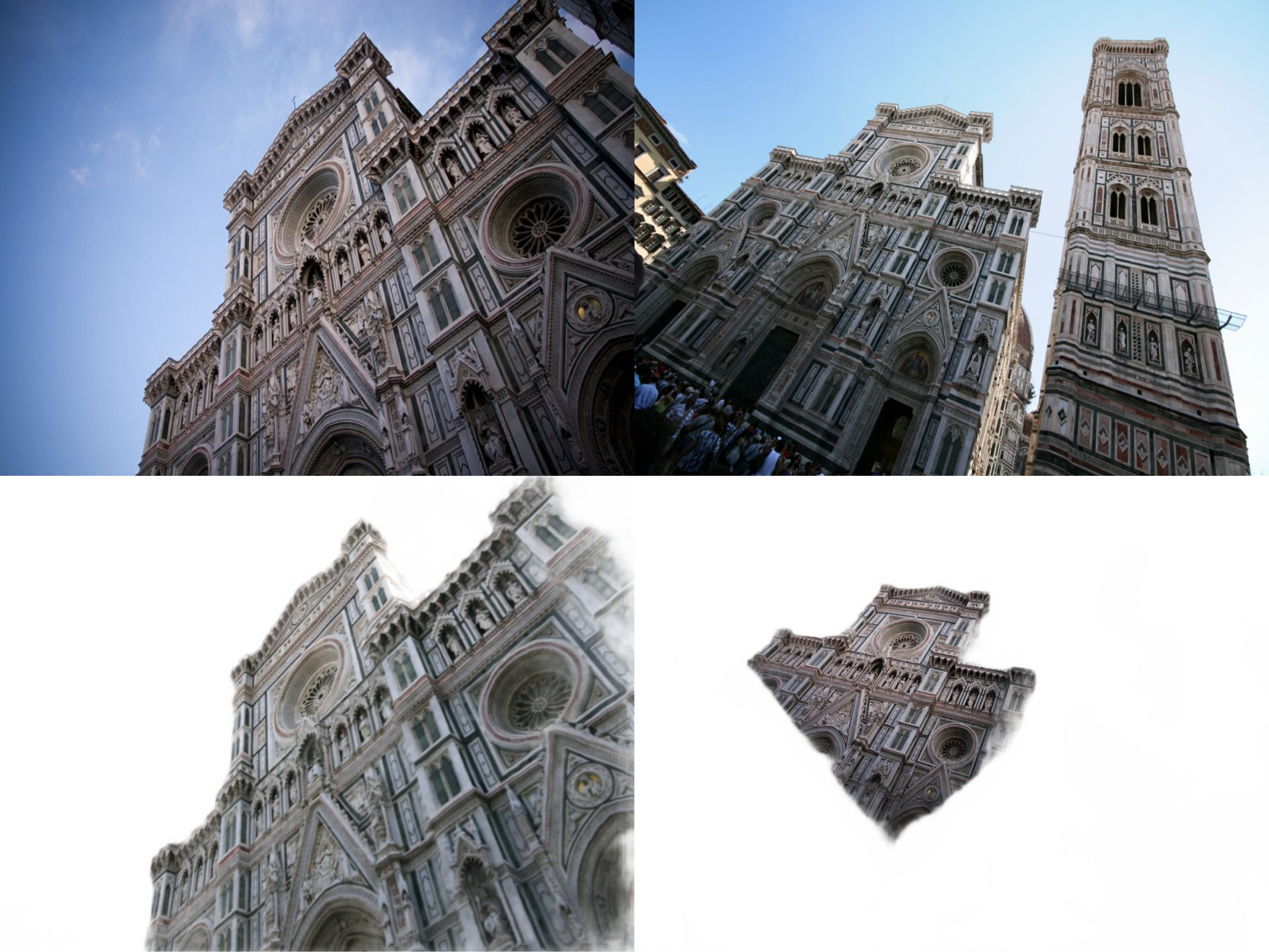}
    \caption{Qualitative example of pair in Florence Cathedral (0032) with DKM warp and certainty.}
    \label{fig:mega0032}
\end{figure}

\begin{figure}
    \centering
    \includegraphics[width=\linewidth]{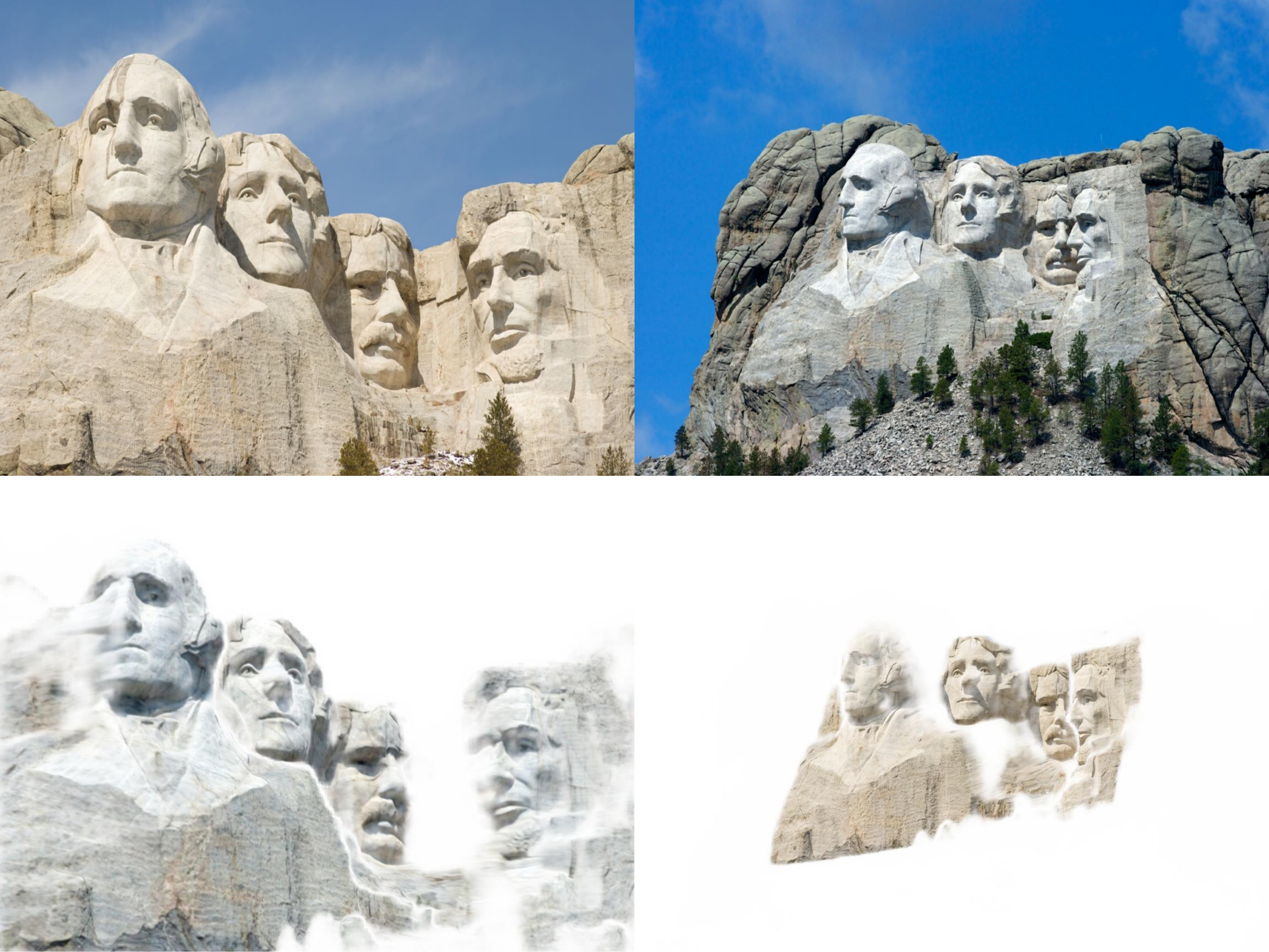}
    \caption{Qualitative example of pair in Mount Rushmore (1589) with DKM warp and certainty.}
    \label{fig:mega1589}
\end{figure}

Since the MegaDepth-1500 benchmark is sampled from only 2 scenes, it is of interest to ascertain that results hold in a wider setting. We therefore sample a total of 1600 pairs from 8 different scenes:
\begin{enumerate}
    \item Piazza San Marco (0008): Example in Figure~\ref{fig:mega0008}.
    \item Sagrada Familia (0019): Example in Figure~\ref{fig:mega0019}.
    \item Lincoln Memorial Statue (0021): Example in Figure~\ref{fig:certainty}.
    \item British Museum (0024): Example in Figure~\ref{fig:mega0024}.
    \item Tower of London (0025): Example in Figure~\ref{fig:method}.
    \item Florence Cathedral (0032): Example in Figure~\ref{fig:mega0032}.
    \item Milan Cathedral (0063): Example in Figure~\ref{fig:qualitative}.
    \item Mount Rushmore (1589): Example in Figure~\ref{fig:mega1589}.
    
\end{enumerate}
We use the same protocol as in MegaDepth-1500. We call this new benchmark \emph{MegaDepth-8-Scenes}.
Results on this benchmark are presented in Table~\ref{tab:megadepth-8-scenes}. We achieve state-of-the-art results here as well, with a relative performance increase of +3.3 AUC$@5^{\circ}$ compared to the previous best sparse method, and by +8.7 percentage points compared to the previous best dense method.

\begin{table}
\small
    \centering
    \caption{ Pose estimation results on the Megadepth-8-Scenes benchmark, measured in AUC (higher is better). Top section, sparse methods, bottom section, dense methods.}
    \begin{tabular}{l lll}
    \toprule
     Method $\downarrow$\quad\quad\quad AUC $\rightarrow$& $@5^{\circ}$&$@10^{\circ}$&$@20^{\circ}$\\
     \midrule
         ASpanFormer~\cite{chen2022aspanformer}~\tiny{ECCV'22} & 57.2  & 72.1 & 82.9\\
         \midrule
         PDCNet+~\cite{truong2021pdc}~\tiny{Arxiv'21} & 51.8 & 66.6 & 77.2 \\
         \textbf{DKM}  &  \textbf{60.5} & \textbf{74.5} & \textbf{84.2}\\

    \bottomrule

    \end{tabular}

    \label{tab:megadepth-8-scenes}
\end{table}

\subsection{St.~Paul's Cathedral}
\label{sec:st-pauls}
\begin{figure}
    \centering
    \includegraphics[width=\linewidth]{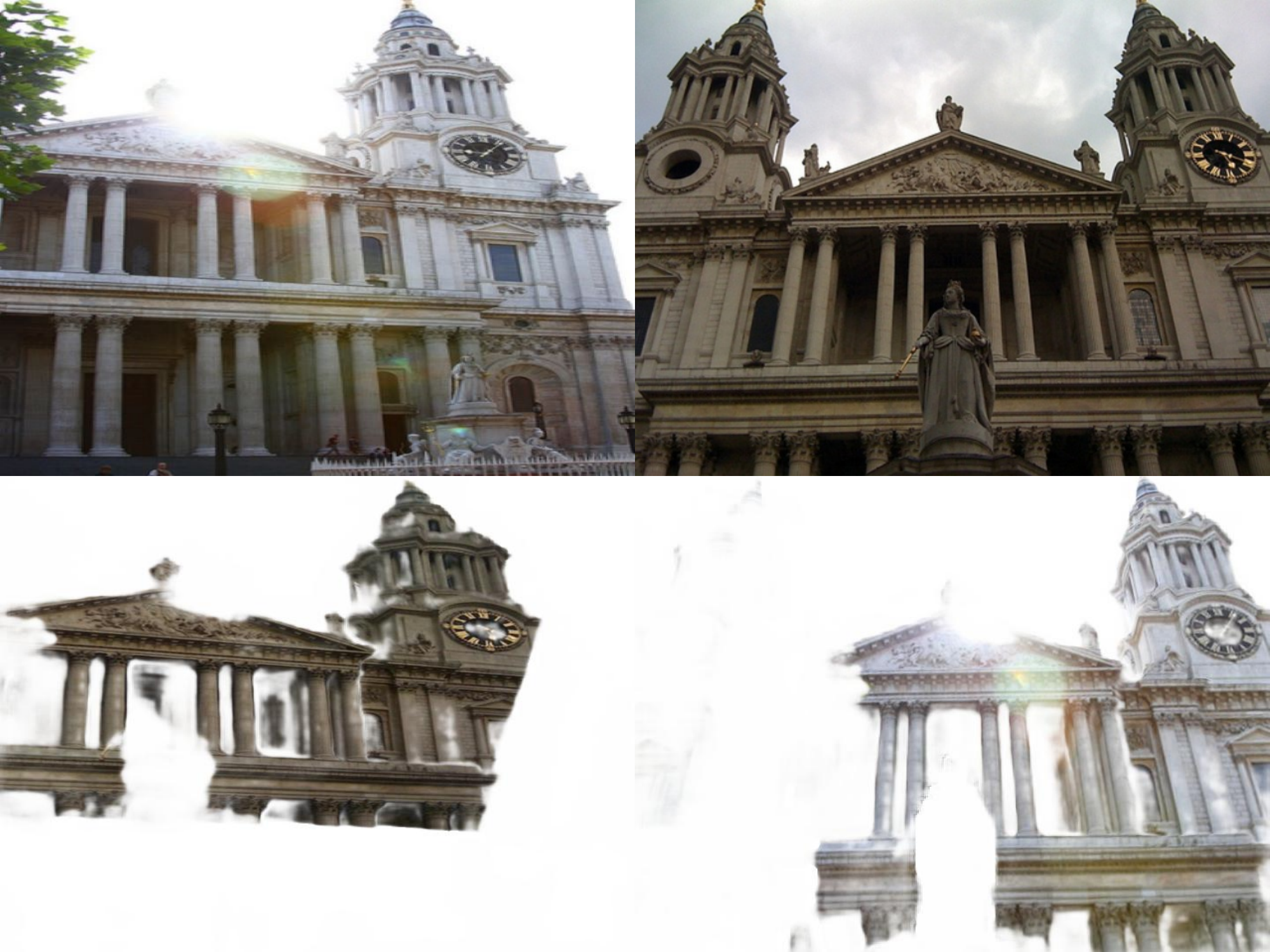}
    \caption{Qualitative example of DKM warp and certainty on the St.~Paul's Cathedral benchmark.}
    \label{fig:st-pauls}
\end{figure}

COTR and ECO-TR~\cite{jiang2021cotr,tan2022eco} are two recent dense methods based on transformer architectures. Here we compare results of our approach compared to those works on the St.~Paul's Cathedral scene. We use the evaluation protocol of ECO-TR. We present results in Table~\ref{tab:st_pauls_cathedral}. We find that our method outperforms both COTR and ECO-TR, achieving a performance increase of +8.0$\text{ mAA}@5^{\circ}$. We additionally present a representative qualitative example in Figure~\ref{fig:st-pauls}.
\begin{table}
\small
    \centering
    \caption{ Pose estimation results on the St.~Paul's Cathedral benchmark, measured in mAA (higher is better). We report the average and estimated standard deviation over five runs.}
    \begin{tabular}{l ll}
    \toprule
     Method $\downarrow$\quad\quad\quad mAA $\rightarrow$& $@5^{\circ}$&$@10^{\circ}$\\
     \midrule
         COTR~\cite{jiang2021cotr}~\tiny{ICCV'21} & 44.3 & 66.0\\
         ECO-TR~\cite{tan2022eco}~\tiny{ECCV'22} & 45.3 & 66.1 \\
         \textbf{DKM}  &  \textbf{53.3} & \textbf{72.1}\\
    \bottomrule

    \end{tabular}

    \label{tab:st_pauls_cathedral}
\end{table}

\section{Further Qualitative Examples}
\subsection{MegaDepth-1500}
\begin{figure}[h!]
    \centering
    \includegraphics[width=\linewidth]{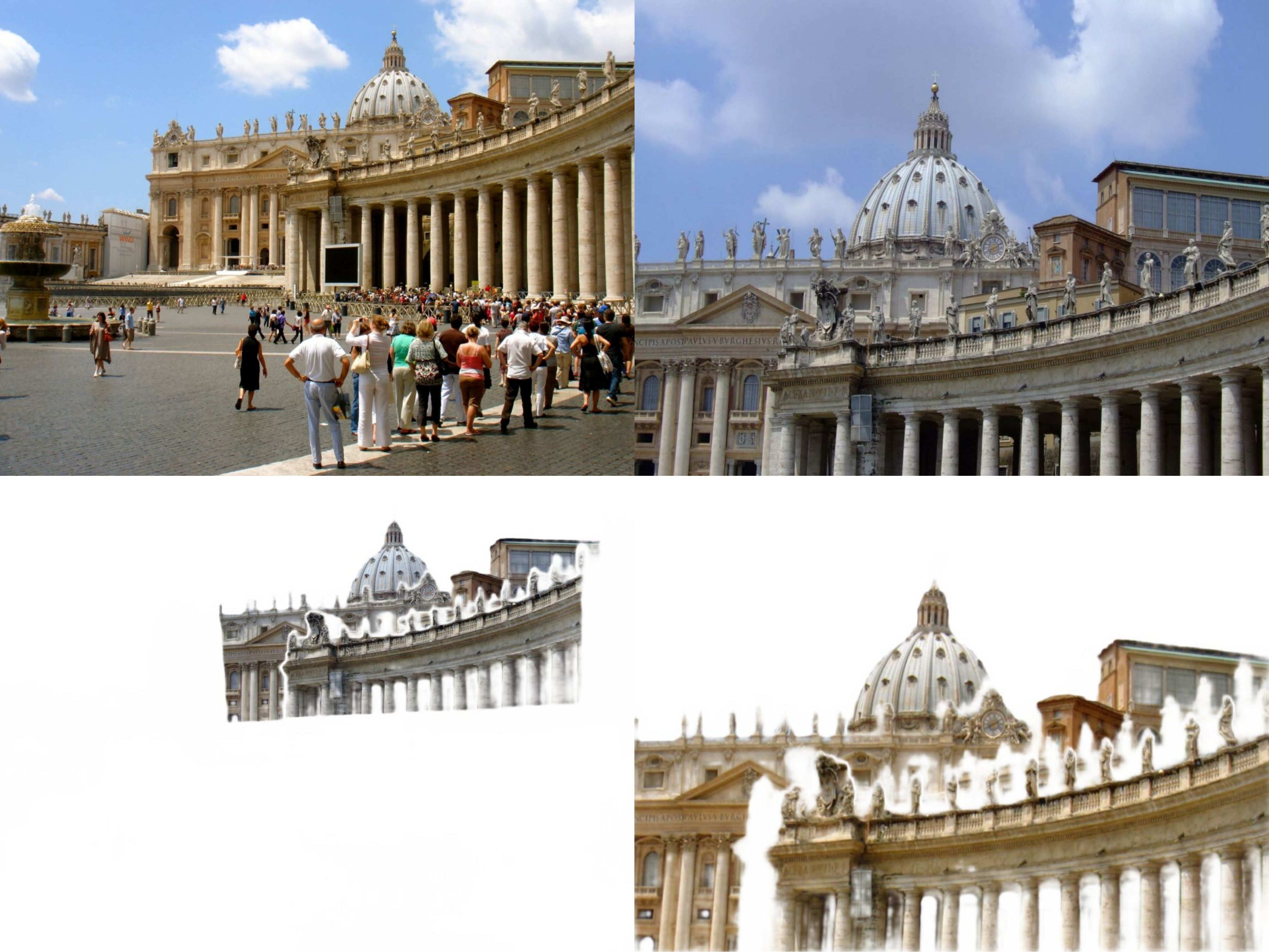}
    \caption{DKM warp and certainty on a pair from the St.\ Peter's Basilica (0015) scene.}
    \label{fig:mega0015}
\end{figure}

In Figure~\ref{fig:mega0015} we present a qualitative example on the St.\ Peter's Basilica (0015) scene.

\subsection{HPatches}
\begin{figure}[h!]
    \centering
    \includegraphics[width=\linewidth]{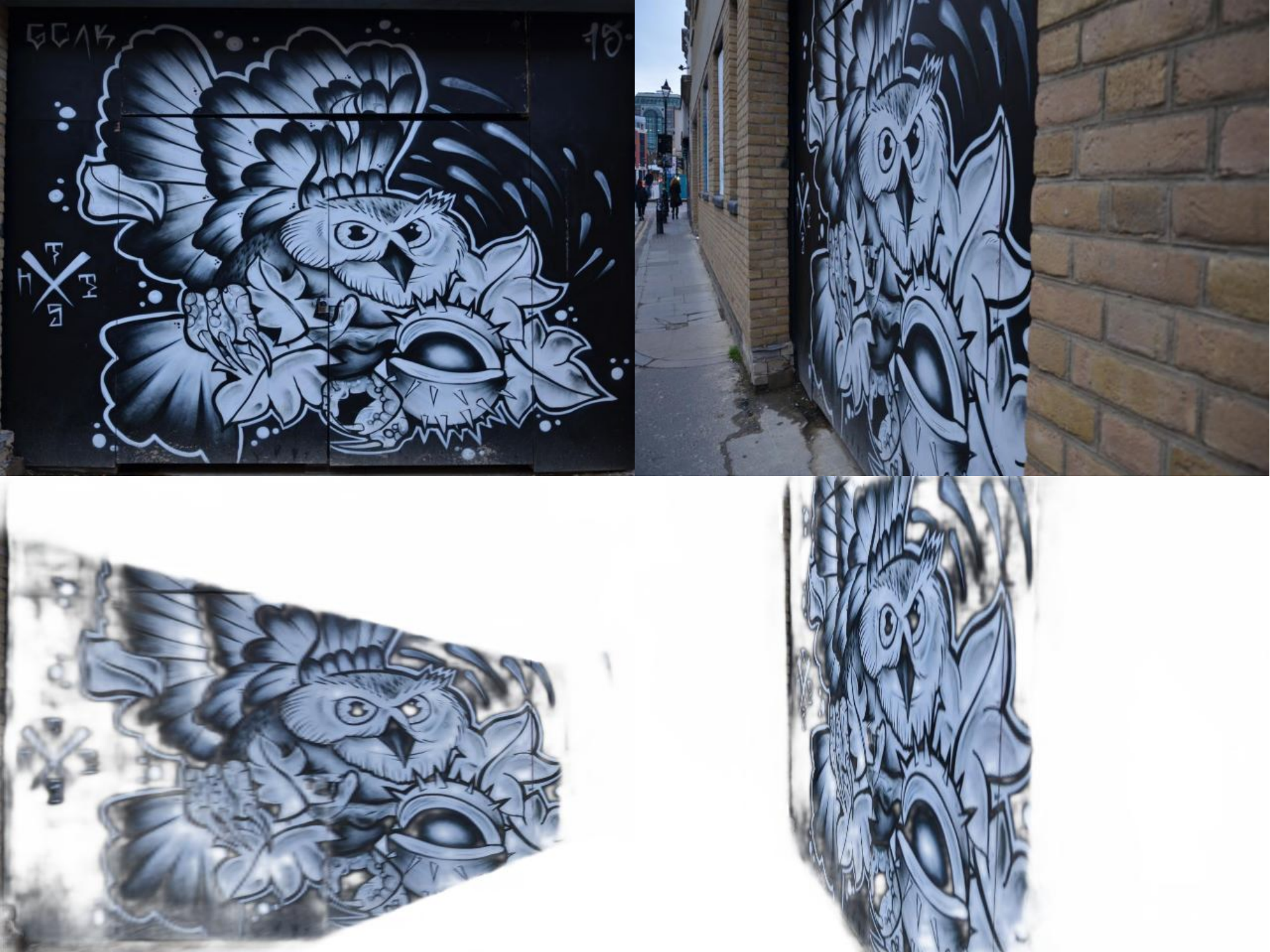}
    \caption{DKM result on the HPatches planar scene v\_bird.}
    \label{fig:hpatches1}
\end{figure}

\begin{figure}[h!]
    \centering
    \includegraphics[width=\linewidth]{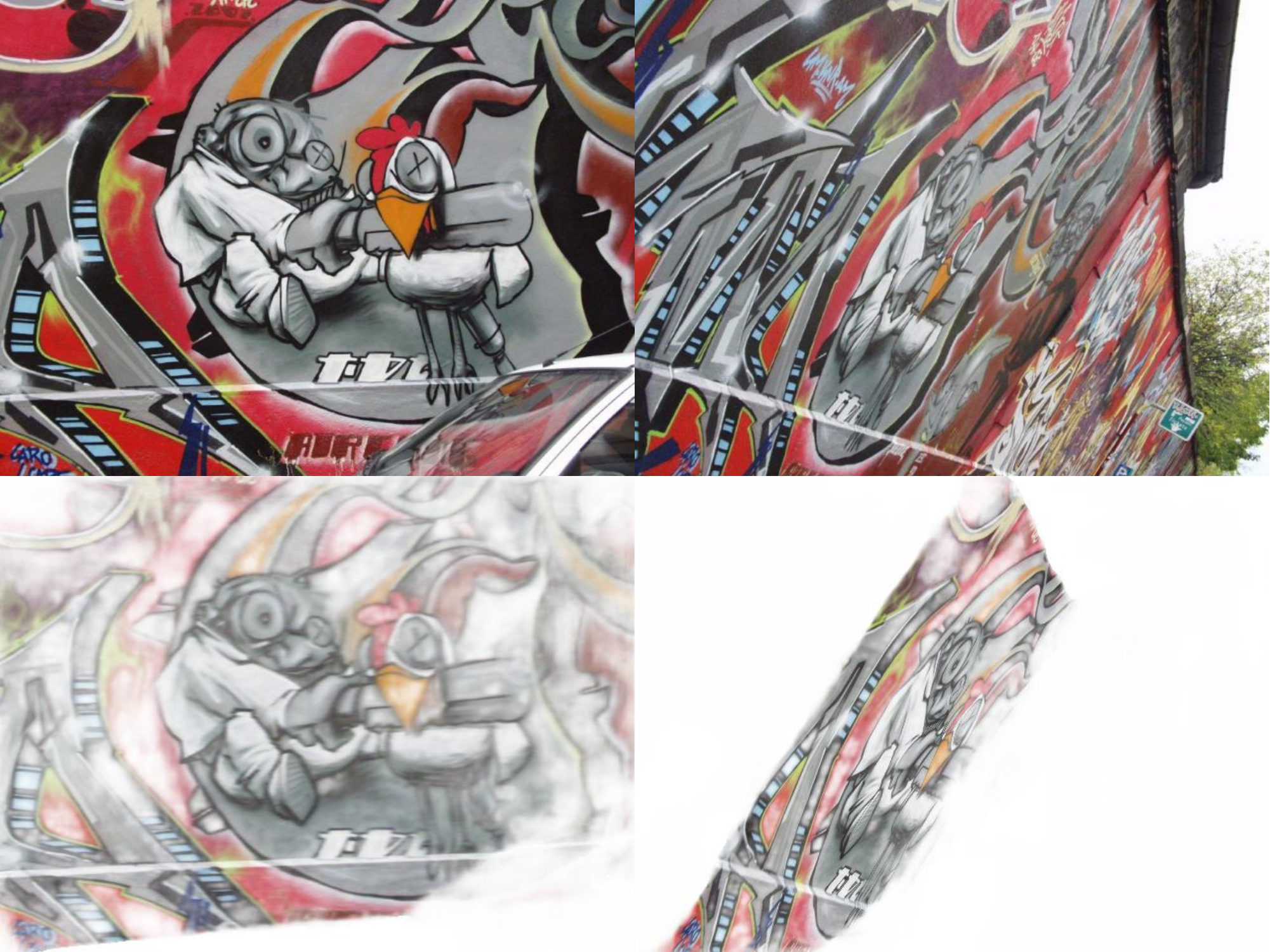}
    \caption{DKM result on the HPatches planar scene v\_graffiti.}
    \label{fig:hpatches2}
\end{figure}

In Figures~\ref{fig:hpatches1} and \ref{fig:hpatches2} we present qualitative results on HPatches. We find that despite not being trained for planar scenes, DKM performs very well here as well.
\subsection{ScanNet}
\begin{figure}
    \centering
    \includegraphics[width=\linewidth]{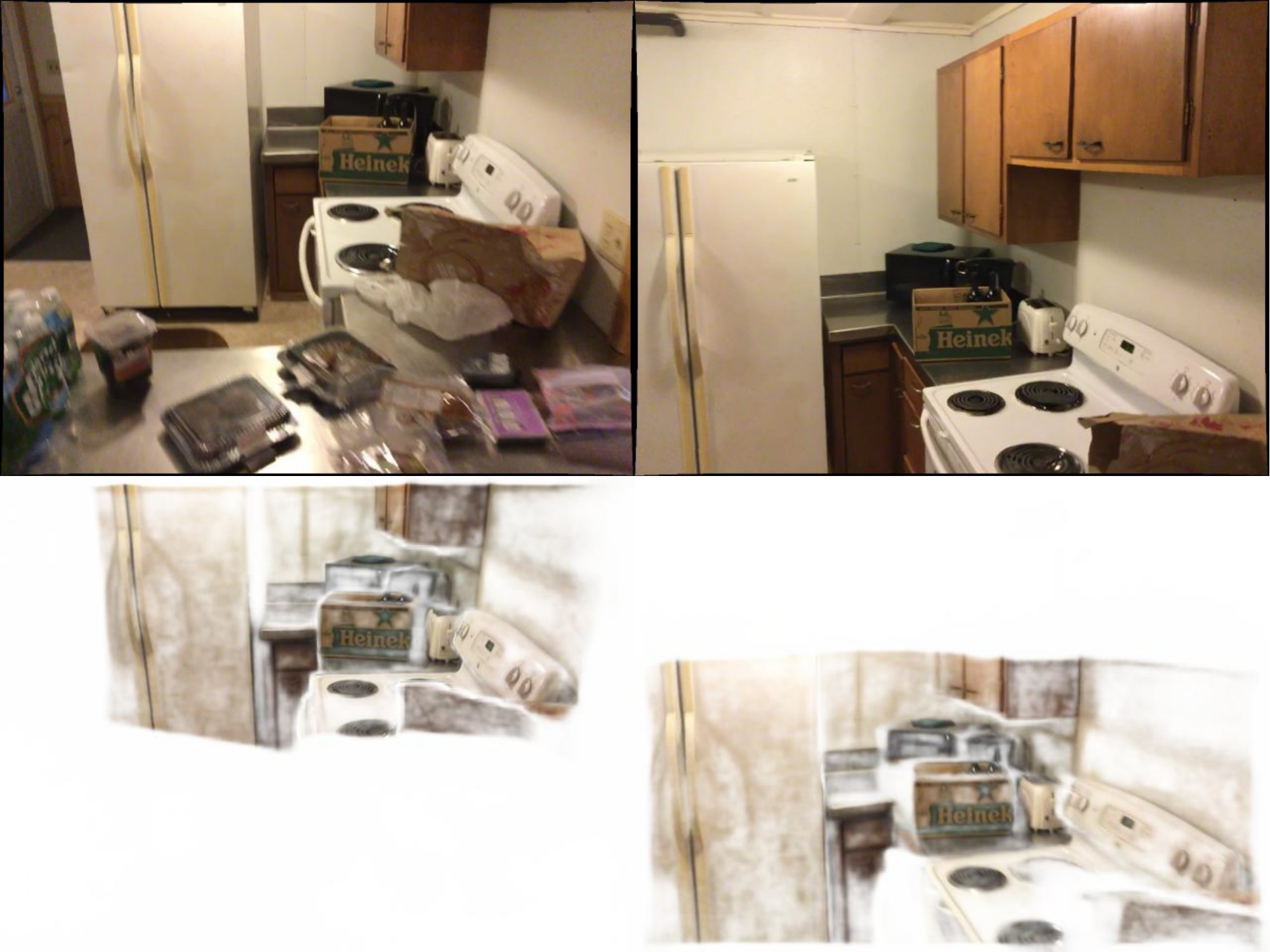}
    \caption{DKM indoor model results on a kitchen scene in the ScanNet-1500 benchmark.}
    \label{fig:scannet}
\end{figure}

In Figure~\ref{fig:scannet}, we present a qualiative example of the indoor model of DKM on the ScanNet-1500 benchmark.

\section{Additional Failure Cases}
\begin{figure}
    \centering
    \includegraphics[width=\linewidth]{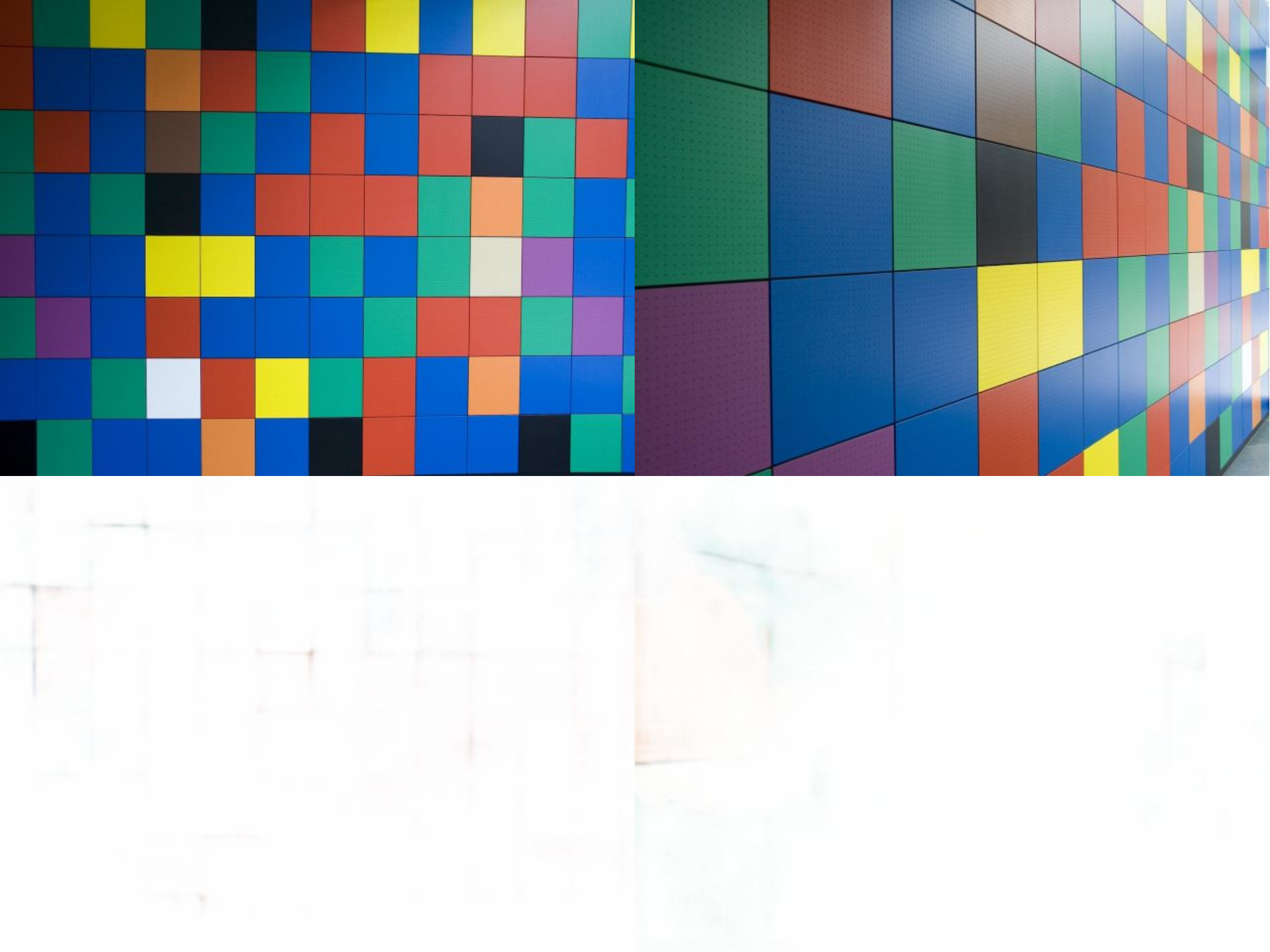}
    \caption{Failure case of DKM. The warp completely fails, and the estimated certainty is very low.}
    \label{fig:hpatches3}
\end{figure}
\parsection{Extreme Lack of Texture}
In Figure~\ref{fig:hpatches3} we show a failure case where our method completely fails. We believe this failure is due to the complete lack of unique local textures. However, the matching is not ill-defined as unique global patterns exist. Encouragingly however, the model predicts a very low certainty for this pair, indicating a well calibrated uncertainty estimate.

\end{document}